\documentclass[lettersize,journal]{IEEEtran}

\usepackage{graphicx}
\usepackage{amsmath,amsfonts}
\usepackage{algorithmic}
\usepackage{algorithm}
\usepackage{array}
\usepackage{booktabs}
\usepackage{textcomp}
\usepackage{stfloats}
\usepackage{url}
\usepackage{verbatim}
\usepackage{graphicx}
\usepackage{cite}
\usepackage{pifont}
\usepackage{xcolor}
\usepackage{caption}
\usepackage{subcaption}
\usepackage{hyperref}
\newcommand{\etal}{\textit{et al.}}

\begin{document}

\title{Component-Aware Sketch-to-Image Generation Using Self-Attention Encoding and Coordinate-Preserving Fusion}
\author{Ali Zia, Muhammad Umer Ramzan, Usman Ali, Muhammad Faheem, Abdelwahed Khamis, Shahnawaz Qureshi


\thanks{Ali Zia, a.zia@latrobe.edu.au, School of Computing, Engineering and Mathematical Sciences La Trobe University. }
\thanks{Muhammad Umer Ramzan, Usman Ali, Muhammad Faheem, \{umer.ramzan, usmanali, mfaheem\}@gift.edu.pk, Gujranwala Institute of Future Technologies(GIFT) University.}
\thanks{Abdelwahed Khamis, abdelwahed.khamis@data61.csiro.au, Data61, CSIRO.}
\thanks{Shahnawaz qureshi, Shahnawaz.qureshi@paf-iast.edu.pk, Sino-Pak Centre for Artificial Intelligence Pak-Austria Fachhochschule Institute of Applied Sciences and Technology.}}



\maketitle

\begin{abstract}

Translating freehand sketches into photorealistic images remains a fundamental challenge in image synthesis, particularly due to the abstract, sparse, and stylistically diverse nature of sketches. Existing approaches, including GAN-based and diffusion-based models, often struggle to reconstruct fine-grained details, maintain spatial alignment, or adapt across different sketch domains. In this paper, we propose a component-aware, self-refining framework for sketch-to-image generation that addresses these challenges through a novel two-stage architecture. A Self-Attention-based Autoencoder Network (SA2N) first captures localised semantic and structural features from component-wise sketch regions, while a Coordinate-Preserving Gated Fusion (CGF) module integrates these into a coherent spatial layout. Finally, a Spatially Adaptive Refinement Revisor (SARR), built on a modified StyleGAN2 backbone, enhances realism and consistency through iterative refinement guided by spatial context. Extensive experiments across both facial (CelebAMask-HQ, CUFSF) and non-facial (Sketchy, ChairsV2, ShoesV2) datasets demonstrate the robustness and generalizability of our method. The proposed framework consistently outperforms state-of-the-art GAN and diffusion models, achieving significant gains in image fidelity, semantic accuracy, and perceptual quality. On CelebAMask-HQ, our model improves over prior methods by 21\% (FID), 58\% (IS), 41\% (KID), and 20\% (SSIM). These results, along with higher efficiency and visual coherence across diverse domains, position our approach as a strong candidate for applications in forensics, digital art restoration, and general sketch-based image synthesis.
\end{abstract}


\section{Introduction}
\label{sec:intro}
Translating hand-drawn sketches into photorealistic images is a long-standing problem in computer vision with practical applications in forensic analysis, digital restoration, and content creation. In particular, face sketch-photo translation is particularly challenging and significant for tasks such as criminal identification \cite{More2023SketchBasedIR}. The fundamental difficulty lies in the substantial domain differences between sketches and photos, which complicates the preservation of semantic consistency during translation \cite{cohen2018distribution}. Therefore, converting a sketch into a realistic photo is a challenging task in computer vision. Sketches are abstract and simple, often lacking essential details like colour, texture, and shading needed for lifelike images. This difference between sketches and photos requires the model to fill in missing information and generate fine details, which can vary based on the artist’s style. Additionally, sketches differ greatly in quality, detail, and focus depending on who draws them, making it harder for the model to learn consistent patterns. Moreover, Sketches, often created under nonideal conditions by artists, frequently lack essential details and exhibit misaligned features, leading to inadequate representation of facial characteristics. Traditional approaches often struggle with preserving local fine-grain details and maintaining semantic consistency, which are crucial for accurately reconstructing facial features from sketches. To address these challenges, Generative Adversarial Networks (GANs) have played a key role, allowing accurate manipulation of facial features through integrated encoder-decoder architectures \cite{Wu2023SketchSceneSS}. 

While diffusion-based methods, such as Stable Diffusion, have achieved remarkable results in general-purpose image generation, they rely on iterative noise sampling, making them computationally expensive and difficult to fine-tune for domain-specific tasks like face sketch-to-image translation~\cite{rombach2022high}. Diffusion models, while effective for image generation tasks like sketch-to-photo translation, exhibit several weaknesses related to the quality of generated images. One major issue is their difficulty in producing fine-grained details, such as realistic textures or intricate facial features, especially when starting from sparse inputs like sketches, often resulting in blurry or incomplete outputs \cite{dhariwal2021diffusion}. Additionally, these methods often fail in scenarios requiring precise spatial alignment and high-frequency detail reconstruction \cite{koley2024s}. To address this, deterministic generative models provide more robust solutions by preserving geometric coherence and facial structure throughout the synthesis process.

Conditional Generative Adversarial Networks (cGANs) improve generative models by incorporating contextual information, refining image quality and preserving facial identity~\cite{Hu2020FacialAS}. Additionally, Deep Convolutional GANs (DCGANs) have enhanced training stability and visual realism in synthesised images~\cite{mathew2021deep,chen2020deepfacedrawing,ramzan2024locally}. However, persistent issues such as information loss and blurring remain, often caused by poor manifold alignment between the sketch and photo domains~\cite{bi2021face}.

Despite significant advances, current sketch-to-image generation methods exhibit persistent limitations. GAN-based models often fail to preserve fine-grained, identity-relevant details and struggle with local semantic consistency due to limited spatial attention mechanisms and holistic processing of sketches \cite{bi2021face, Hu2020FacialAS}. Diffusion-based models, though strong in general-purpose synthesis, are computationally expensive and prone to producing blurry or structurally inconsistent outputs, especially when handling sparse, stylised, or misaligned sketches \cite{dhariwal2021diffusion, koley2024s}. Moreover, few approaches explicitly model component-level structure or spatial alignment, limiting their generalizability across facial and non-facial domains. These gaps highlight the need for a framework that can (1) model localised features, (2) preserve spatial coherence, and (3) adaptively refine visual quality in both structured and unstructured sketch domains.

To address these challenges, we propose a component-aware, self-refining framework for sketch-to-image generation, structured around a novel two-stage learning architecture. In the first stage, localised semantic features are extracted through a self-attention-based encoder to capture fine-grained component-level information. In the second stage, these features are spatially integrated into a coherent global image, followed by an adaptive refinement process that enhances realism, texture quality, and structural fidelity. The proposed framework is designed to balance \textbf{semantic precision}, \textbf{spatial alignment}, and \textbf{perceptual quality}. Our framework demonstrates strong generalisation across both facial and non-facial sketch domains. Our main contributions are:

\begin{itemize}
    \item \textbf{Component-aware feature encoding via self-attention autoencoders}, enabling localized semantic representation of sketch regions (e.g., eyes, nose, mouth). This directly addresses the lack of spatial disentanglement and fine-grained detail modeling in prior GAN-based approaches.
    
    \item \textbf{Coordinate-Preserving Gated Fusion (CGF) for spatially consistent synthesis}, which mitigates information loss and geometric distortion by maintaining alignment across semantically decomposed regions, overcoming the spatial misalignment seen in both GANs and diffusion-based models.
     
    \item \textbf{Spatially Adaptive Refinement Revisor (SARR)} based on a modified StyleGAN2 architecture, which enhances image realism, identity preservation, and high-frequency detail through iterative refinement, providing an efficient alternative to costly diffusion refiners.
\end{itemize}

We develop our methodology from a facial data perspective, as face sketch-to-image translation represents one of the most semantically complex and practically critical challenges in the domain. Tasks such as forensic reconstruction and identity recovery require high fidelity, structural accuracy, and preservation of fine-grained features, making the facial domain a rigorous and representative foundation for architectural design.

While our framework is optimised for facial structure through component-aware modelling and adaptive refinement, we demonstrate strong generalisation to non-facial sketches, including objects with diverse topologies such as shoes and chairs. Extensive evaluations across multiple benchmarks show that our method consistently outperforms state-of-the-art GAN and diffusion-based approaches in terms of image fidelity, semantic coherence, and perceptual realism. These findings underscore the strength of modelling localised semantics and spatial relationships in advancing sketch-to-image synthesis.

\section{Related Work}
\label{sec:Related Work}

The literature on sketch-to-image translation has seen various approaches leveraging advanced neural architectures. In particular, More \etal~\cite{More2023SketchBasedIR} explored the use of deep convolutional neural networks (DCNN) to transform sketches into realistic images. General translation challenges have been addressed through various methodologies aimed at increasing the realism of input sketches \cite{chen2018sketchygan}. A significant development in this area has been the pix2pix framework introduced by \cite{isola2017image}, which employs a U-Net-based cGAN. Despite its innovative approach, pix2pix often produces low-resolution images (256x256), affecting output quality. This limitation was addressed by \cite{wang2018high} with Pix2PixHD, which enhances image quality by splitting the generator into two sub-networks trained on high-resolution images.

Although these methods have successfully generated visually impressive images from sketches, they often fail to capture accurate structural details, such as specific facial features. To improve realism, \cite{li2019linestofacephoto} introduced the LinesToFacePhoto framework, which uses a conditional self-attention mechanism within a generative adversarial network to better generate face images from line sketches.

Further refinements include the work of \cite{zhao2019generating}, who developed methods to control specific facial attributes such as gender and age in the generated images. Meanwhile, Hicsonmez \etal~\cite{hicsonmez2023improving} proposed a technique to improve the colourisation accuracy of sketches using adversarial segmentation loss, addressing misalignments between colourized outputs and their corresponding sketches. Moreover, \cite{richardson2021encoding} introduced pixel2style2pixel (pSp), an image translation framework that maps images into a latent space for generating stylized outputs. Ramzan \etal~\cite{ramzan2024locally} proposed a two-stage framework combining a Convolutional Block Attention-based Auto-encoder,  but their approach remains limited to facial sketches and lacks self-attention based component-wise modelling, coordinate-preserving fusion, and spatially adaptive refinement, all of which are crucial for capturing fine-grained structural details, improving spatial alignment, and enabling generalisation to non-facial domains. Recent advancements by Guo \etal~ \cite{guo2024image} with the Offset-based Multi-Scale Codes GAN Encoder (OME) integrate a multi-scale latent codes GAN encoder with iterative update synthesis, improving detailed and realistic image generation. However, the reliance on pre-trained GANs limits the application of OME in precise sketch-based image synthesis tasks. Building upon existing research, our work addresses the persistent challenge of achieving high fidelity and structural accuracy in sketch-to-image translation, a gap highlighted by prior studies. 

Diffusion models have recently advanced sketch-to-image generation by operating in latent space for high-fidelity synthesis. ControlNet\cite{zhang2023adding} augments frozen T2I models with control branches to enforce structural guidance from sketches, while T2I-Adapter\cite{mou2024t2i} uses lightweight modules to align external controls with internal features, enabling composable conditioning. Another approach, such as SDEdit\cite{meng2021sdedit}, leverages user strokes or edge maps to refine structural coherence. Despite strong results, limitations include reliance on large pretrained datasets, sensitivity to sparse or noisy sketches, conflicts in multi-control settings, and high computational costs compared to GANs.



\begin{figure*}[t]
\centering
\includegraphics[width=1\textwidth]{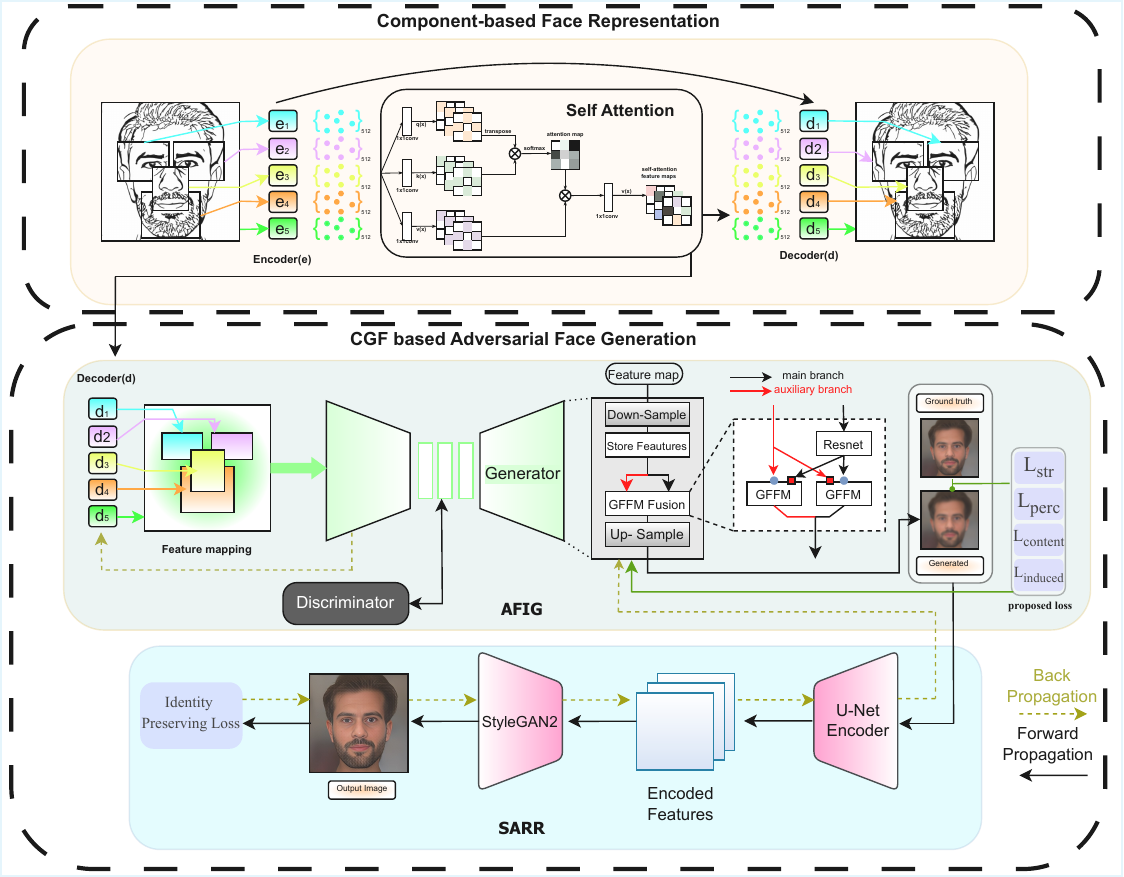}
\caption{Illustration 
of the proposed sketch to image generation architecture. (Top) \textbf{Component-based Face Representation Learning}. The self-attention mechanism is applied to each component of the sketch face to refine feature representations in the autoencoder. (Bottom) \textbf{CGF based Adversarial Face Generation}. The feature descriptors of each facial component are converted into feature maps, and these feature maps undergo the training process in AFIG module. The bidirectional arrow connecting the discriminator represents joint fine-tuning of the trained component encoders during the second training stage. Finally, the generated image passed through a SARR module to fine-tune the details and quality of the generated image using identity-preserving loss. 
}
\label{fig:architecture}
\end{figure*}

\section{Self-Improving Sketch to Image Face Generation Framework}\label{sec:method}

Two-stage architectures have been widely adopted in sketch-to-image translation due to their ability to progressively refine structural representations and enhance visual realism\cite{lee2022two,kim2023dcface}. Building on this paradigm, we design a framework that introduces targeted improvements to both stages, thereby addressing limitations of prior approaches in preserving local details and maintaining semantic consistency. As illustrated in Fig. 1, the first stage performs component-based face representation learning, where a self-attention mechanism is incorporated to mitigate embedding discontinuities across local facial regions. The second stage employs a Coordinate-preserving Gated Fusion (CGF) module for adversarial generation, with the Spatially Adaptive Refinement Revisor (SARR) integrated to provide structural feedback and improve the fidelity of cGAN outputs.


\subsection{Component-based Face Representation Learning}\label{first training stage}
In this stage, we decompose the input facial sketch into five distinct components, \textit{left eye, right eye, nose, mouth, and remaining facial features}, to enhance localised feature representation. Each component is processed by an independent autoencoder, denoted as \( \{E_c, D_c\} \), where \( E_c \) extracts latent embeddings and \( D_c \) reconstructs the corresponding region. This decomposition allows for fine-grained feature learning, which is particularly beneficial in sketch-based generation, where crucial details such as eye structure and mouth shape are often distorted or missing.

To maintain consistency in feature space, all autoencoders share a common latent dimension of $512$, enforced through a fully connected layer. This ensures that spatially distinct features are mapped into a unified representation, facilitating smooth integration in later stages of image synthesis.

Unlike previous methods that manually construct smooth manifolds for facial component alignment~\cite{chen2020deepfacedrawing}, our approach automatically learns spatial dependencies using a self-attention mechanism. Specifically, we integrate the self-attention framework proposed within the encoder, allowing the network to dynamically capture contextual relationships between facial regions. This mechanism improves feature coherence, ensuring that generated facial components align naturally while preserving global facial structure. By leveraging self-attention, our network simultaneously learns both localised feature embeddings and global structural dependencies, addressing key challenges in sketch-to-image translation.

\subsection{CGF-based Adversarial Face Generation}\label{second_stage_training}
The second stage of our framework refines the component-based feature representations extracted in the first stage. This is achieved through an Adaptive Feature Integration Generator (AFIG) module and \textit{Spatially Adaptive Refinement Revisor (SARR)} module. 


\subsubsection{Adaptive Feature Integration Generator (AFIG)}
AFIG consists of two core components:
\begin{itemize}
    \item The Feature Mapping (FM) module, which processes encoded facial components and maps them into spatially structured feature representations.
    \item The Coordinate-Preserving Gated Fusion (CGF) module, which refines these mapped features and ensures structural consistency through a spatially aware gating mechanism.
\end{itemize}
This structured processing enables precise facial component integration while preventing distortions that commonly arise in sketch-to-image translation.

\textbf{Feature Mapping (FM) module:}

FM module plays a crucial role in ensuring that facial components retain their structural integrity before integration into the final image synthesis process. Unlike standard feature extraction pipelines, FM employs five independent decoders, each responsible for mapping feature vectors into spatially structured representations. These feature maps retain spatial information, allowing finer control over facial attributes such as eye symmetry and mouth shape. By preserving these spatial relationships, FM facilitates a more natural fusion of localised features, preventing deformations commonly observed in holistic image synthesis approaches.

\textbf{Coordinate-Preserving Gated Fusion (CGF) module:}

At the core of our framework, CGF refines the feature maps produced by FM. The CGF serves as the generator 
and incorporates a dual-branch architecture:
\begin{itemize}
    \item The \textit{main branch} utilises residual blocks for high-level feature propagation.
    \item The \textit{auxiliary branch} transfers coarse feature maps from early layers to later network stages, mitigating the risk of vanishing details.
\end{itemize}
This dual-branch setup ensures efficient information flow and facilitates high-fidelity sketch-to-image translation.

Unlike standard feature fusion mechanisms, CGF preserves spatial consistency by integrating a \textit{Spatial-Preserving Convolution (SPConv)} mechanism, which ensures that fine-grained details are maintained across the generated image. 

Traditional feature fusion methods often rely on concatenation or additive blending, which can lead to feature misalignment, loss of fine-grained details, and spatial inconsistencies. To address this, we introduce the \textit{Coordinate-Preserving Gated Fusion (CGF)} module, which dynamically selects and fuses feature maps while preserving spatial integrity.

The gating function \( g(\mathbf{C}) \) is designed to enhance feature selection by assigning adaptive weights to different spatial regions, ensuring that features corresponding to different facial components (e.g., eyes, nose, mouth) remain correctly aligned. This is achieved using a spatially-aware gating mechanism that modulates fusion weights based on the input coordinate map. The fusion process is defined as:
\begin{equation}
    y_{i} = h(\mathbf{x})_i \cdot g(\mathbf{C}_i),
\end{equation}

where:
\begin{itemize}
    \item \( h(\mathbf{x}) \) applies a \textit{Spatial-Preserving Convolution (SPConv)} to extract localized features from the input sketch.
    \item \( g(\mathbf{C}) \) generates a \textit{gating mask} from the static coordinate map \( \mathbf{C} \in \mathbb{R}^{n_x \times n_y \times 2} \), guiding feature selection and preventing spatial distortion. Here $n_x$ and $ n_y$ denote the spatial dimensions, width and height, of the feature map.
\end{itemize}

Unlike traditional Gated Feature Fusion Modules, CGF ensures that local spatial dependencies are retained, reducing artefacts and improving structural coherence in the synthesised images. The refined feature maps are then passed through a GAN, transforming them into photorealistic outputs.

\subsubsection{Spatially Adaptive Refinement Revisor (SARR)}
While the AFIG module is responsible for initial sketch-to-image translation, it may introduce texture inconsistencies, loss of fine-grained details, or identity mismatches in complex facial structures. To address these challenges, we employ a Spatially Adaptive Refinement Revisor (SARR) that refines image quality using a combination of multi-scale feature processing and spatial adaptation mechanisms.

SARR builds upon the StyleGAN2 architecture, integrating Spatial Feature Transform (SFT) layers to dynamically adjust feature distributions based on the input sketch. Unlike standard refinement approaches, which rely on post-hoc filtering, SARR iteratively corrects artefacts using an adaptive feedback loop, progressively improving visual coherence, making it particularly effective for identity-sensitive applications such as forensic reconstruction and digital restoration.

The SARR module complements AFIG by refining its outputs using a UNet~\cite{ronneberger2015u} along with a modified StyleGAN2~\cite{karras2020analyzing} architecture integrated with Spatial Feature Transform (SFT) layers~\cite{wang2018recovering}.

By utilising a UNet encoder, SARR effectively captures multi-scale features and maintains structural coherence through skip connections, a design particularly suited for the sparse and often abstract nature of sketches. The StyleGAN2-based decoder, known for generating high-quality and realistic imagery, is further enhanced with SFT layers, which enable adaptive modulation of multi-scale features. This enhancement improves fine-grained details such as textures and edges, ensuring high-resolution image fidelity. The combination of global structural accuracy and precise textural refinement results in high-fidelity sketch-to-image translation.

SARR operates within an iterative feedback loop, progressively refining image quality. This allows SARR to correct artefacts and improve structural consistency. Unlike AFIG, 
which primarily focuses on generating high-quality initial outputs, SARR employs a loss function aimed at enhancing identity and style preservation. The \textit{identity loss} \( L_{\text{id}} \) ensures that the refined images retain identity-specific features:
    \begin{equation}
    L_{\text{id}} = \lambda_{\text{id}} \left\| \eta(\hat{y}) - \eta(y) \right\|_1,
    \end{equation}
where \( \hat{y} \) and \( y \) denote the generated and real images, respectively, and \( \eta \) represents a pre-trained ArcFace\cite{deng2018arcface}
model. The weighting factor \( \lambda_{\text{id}} \) balances the contribution of identity preservation during training\cite{wang2021gfpgan}. 

Through its iterative refinement mechanism, SARR excels in enhancing the perceptual and structural fidelity of generated images. When combined with AFIG, the two components create a powerful framework that balances realism, accuracy, and identity preservation across complex facial sketches. This synergy allows the proposed method to outperform both GAN-based and diffusion-based approaches in domain-specific tasks.

\subsubsection{Optimisation and Loss Functions}
To optimise CGF and improve perceptual quality, we employ a combination of loss functions:
\begin{itemize}
    \item \textbf{Pixel-wise \( L_1 \) Loss:} Ensures accurate reconstruction by minimising the absolute difference between generated and real images.
    \begin{equation}
    L_{1} = \frac{1}{N} \sum_{i=1}^{N} \left| G_i - R_i \right|,
    \end{equation}
    where \( G_i \) and \( R_i \) denote the generated and ground truth images.

    \item \textbf{Adversarial Loss:} Adversarial loss introduces realism by encouraging the generator to produce outputs that align with the natural image manifold, thereby improving texture sharpness and reducing blurriness.
    \begin{equation}
    \mathcal{L}_{\text{GAN}} = \mathbb{E}_{y \sim R} [\log d(y)] + \mathbb{E}_{x \sim G} [\log (1 - d(g(x)))],
    \end{equation}
    where \( g \) and \( d \) denote the generator and discriminator.

    \item \textbf{Perceptual Loss:} 
    Perceptual Loss\cite{wang2021gfpgan} ensures high-level feature similarity between generated and real images, which ensures outputs remain close to the ground truth by preserving global structure and perceptual similarity.
    \begin{equation}
    L_{\text{perc}} = \frac{1}{N} \sum_{i=1}^{N} \left\| F(G_i) - F(R_i) \right\|,
    \end{equation}
    where \( F(\cdot) \) represents a pre-trained VGG11 network.

    \item \textbf{Gram Matrix Loss:} 
    Gram Matrix Loss\cite{gatys2016image} utilises Gram matrix correlations between feature maps to encode stationary multi-scale texture patterns, thereby enriching local details while being agnostic to global spatial arrangement, which maintains stylistic consistency across the generated image.
    \begin{equation}
    L_{\text{Gram}} = \frac{1}{L} \sum_{l=1}^{L} \alpha_l \left\lVert M_l(G) - M_l(R) \right\lVert_2,
    \end{equation}
    where \( M_l(G) \) and \( M_l(R) \) are the Gram matrices at layer \( l \) of a pre-trained VGG19 model.
\end{itemize}


This comprehensive optimisation strategy, combined with CGF’s robust feature fusion capabilities and SARR’s iterative refinement, ensures that our framework generates high-quality images with precise \textit{structural alignment, perceptual realism, and identity preservation}.
 
\section{Experiments and Results}\label{sec:results and discussion}
This section outlines the experimental setup, results, and ablation study, demonstrating the framework's robustness and generalisation across various sketch types.

\subsection{Experimental Setting}
\subsubsection{Datasets}\label{subsec:datapreparation}

We train and evaluate our architecture on three benchmark datasets: CelebAMask-HQ \cite{lee2020maskgan}, CUHK \cite{wang2008face}, and CUFSF \cite{zhang2011coupled}. 
Data preprocessing was applied to remove occluded faces, and the super-resolution technique \cite{yu2018face} was applied to obtain high-quality input image data. 
Since the CelebAMask-HQ dataset includes full-body images of celebrities, we adapted these images for the model by cropping face regions using a pre-trained Haar cascade classifier~\cite{viola2001rapid}, which is commonly used for face recognition. After preprocessing, we obtained 11,000, 667, and 188 sketch-image pairs for the CelebAMask-HQ, CUFSF, and CUHK datasets, respectively. To ensure a fair comparison with similar studies, we used a 90:10 training-to-testing ratio in our experiments.

\begin{figure*}[!ht]
\centering

\rotatebox[origin=c]{90}{\parbox[c]{0\textheight}{\centering Input}}
\begin{minipage}[b]{0.14\textwidth}
  \centering
  \makebox[0pt]{(a)}\\
  \fbox{\includegraphics[width=\textwidth]{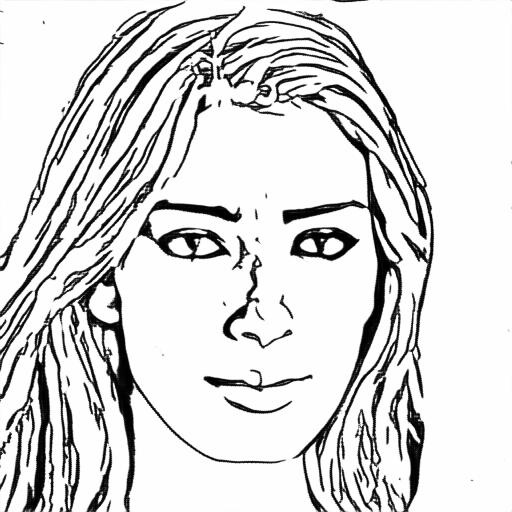}}
\end{minipage}%
\hspace{0.01\textwidth} 
\begin{minipage}[b]{0.14\textwidth}
  \centering
  \makebox[0pt]{(b)}\\
  \fbox{\includegraphics[width=\textwidth]{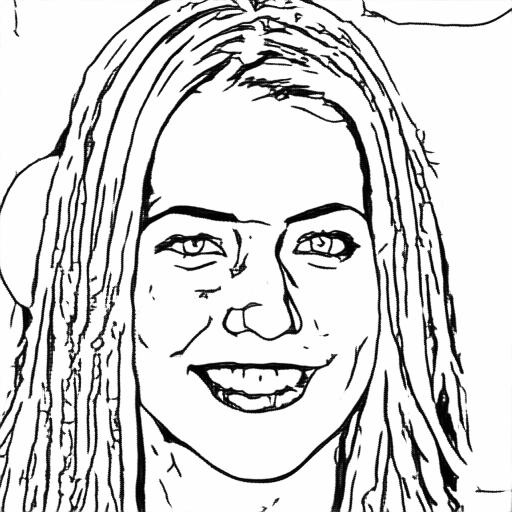}}
\end{minipage}%
\hspace{0.01\textwidth} 
\begin{minipage}[b]{0.14\textwidth}
  \centering
  \makebox[0pt]{(c)}\\
  \fbox{\includegraphics[width=\textwidth]{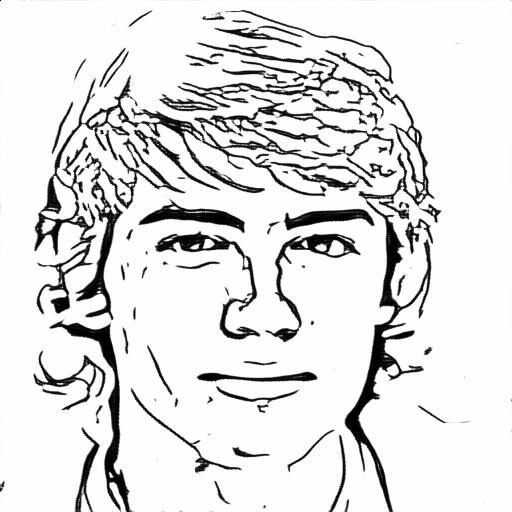}}
\end{minipage}%
\hspace{0.01\textwidth} 
\begin{minipage}[b]{0.14\textwidth}
  \centering
  \makebox[0pt]{(d)}\\
  \fbox{\includegraphics[width=\textwidth]{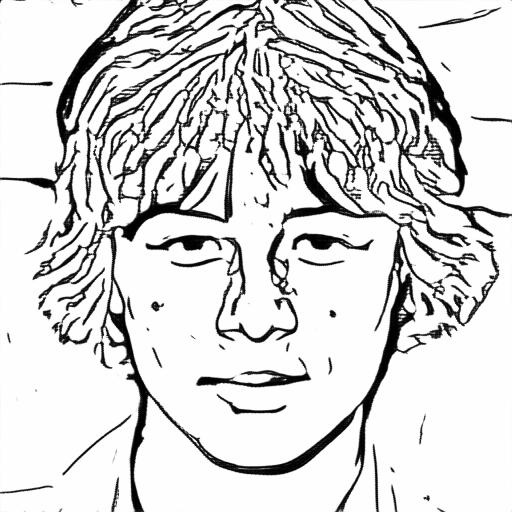}}
\end{minipage}%
\hspace{0.01\textwidth} 
\begin{minipage}[b]{0.14\textwidth}
  \centering
  \makebox[0pt]{(e)}\\
  \fbox{\includegraphics[width=\textwidth]{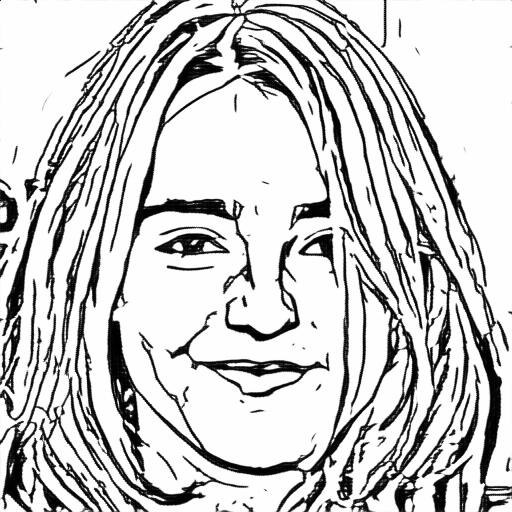}}
\end{minipage}%
\hspace{0.01\textwidth} 
\begin{minipage}[b]{0.14\textwidth}
  \centering
  \makebox[0pt]{(f)}\\
  \fbox{\includegraphics[width=\textwidth]{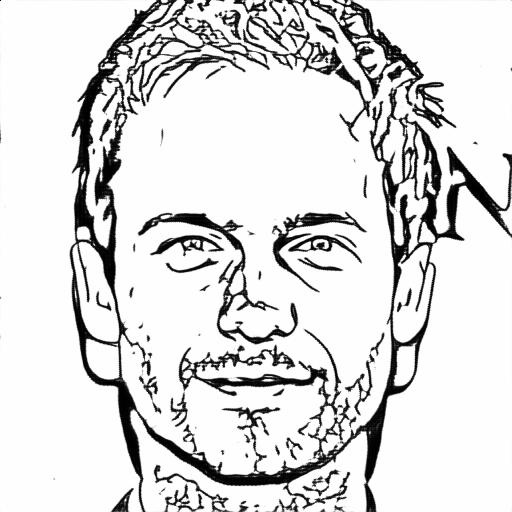}}
\end{minipage}

\vspace{0.5em} 

\rotatebox[origin=c]{90}{\parbox[c]{0\textheight}{\centering GT}}
\begin{minipage}[b]{0.14\textwidth}
  \centering
  \fbox{\includegraphics[width=\textwidth]{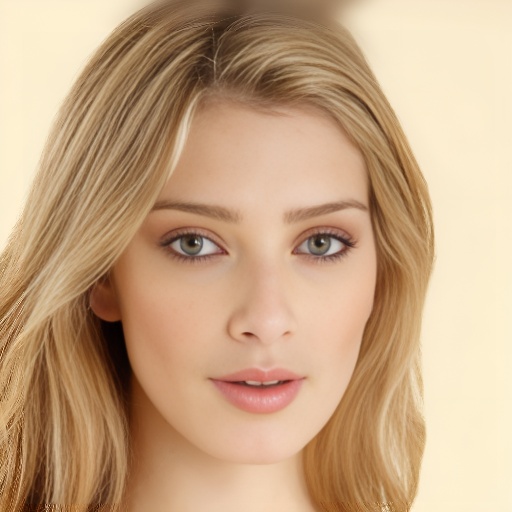}}
\end{minipage}%
\hspace{0.01\textwidth} 
\begin{minipage}[b]{0.14\textwidth}
  \centering
  \fbox{\includegraphics[width=\textwidth]{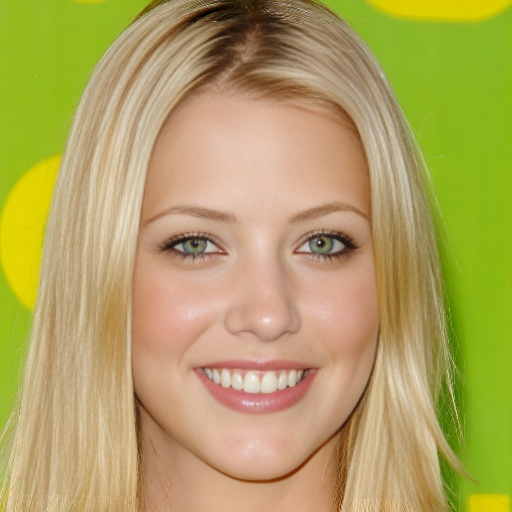}}
\end{minipage}%
\hspace{0.01\textwidth} 
\begin{minipage}[b]{0.14\textwidth}
  \centering
  \fbox{\includegraphics[width=\textwidth]{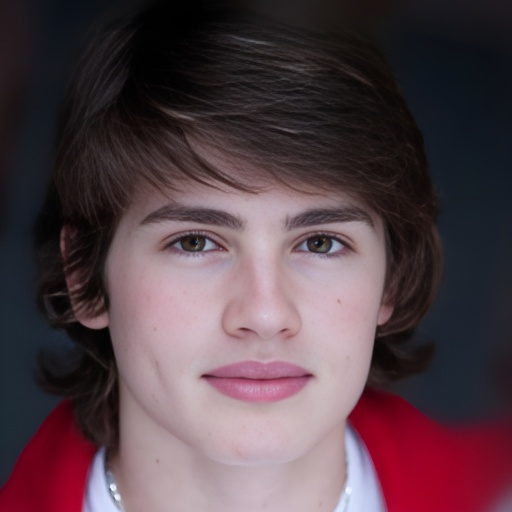}}
\end{minipage}%
\hspace{0.01\textwidth} 
\begin{minipage}[b]{0.14\textwidth}
  \centering
  \fbox{\includegraphics[width=\textwidth]{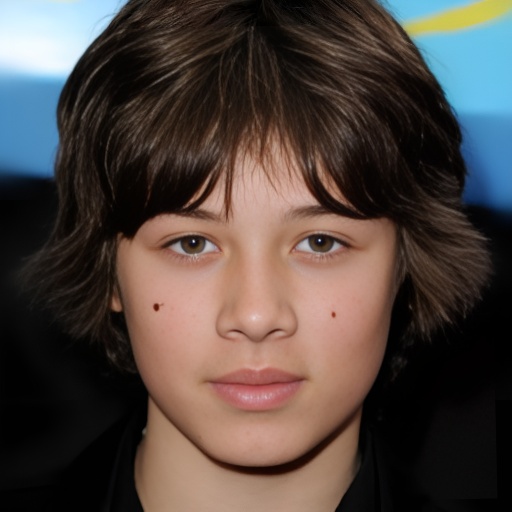}}
\end{minipage}%
\hspace{0.01\textwidth} 
\begin{minipage}[b]{0.14\textwidth}
  \centering
  \fbox{\includegraphics[width=\textwidth]{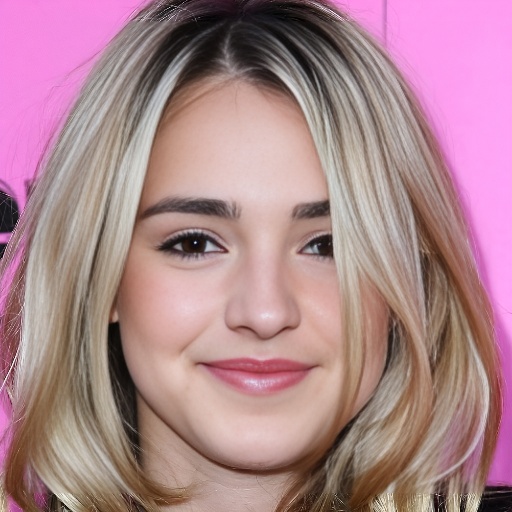}}
\end{minipage}%
\hspace{0.01\textwidth} 
\begin{minipage}[b]{0.14\textwidth}
  \centering
  \fbox{\includegraphics[width=\textwidth]{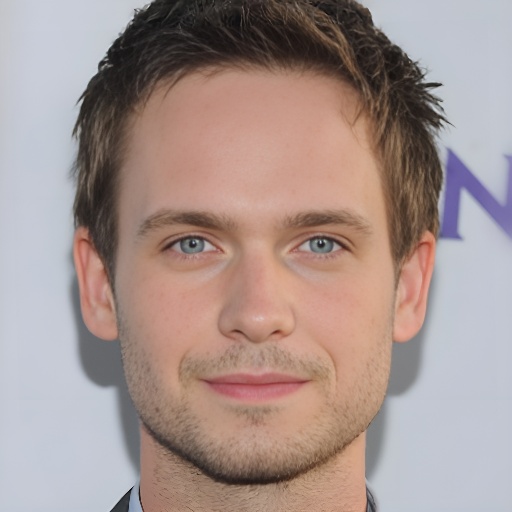}}
\end{minipage}

\vspace{0.5em} 

\rotatebox[origin=c]{90}{\parbox[c]{0\textheight}{\centering Pix2PixHD}}
\begin{minipage}[b]{0.14\textwidth}
  \centering
  \fbox{\includegraphics[width=\textwidth]{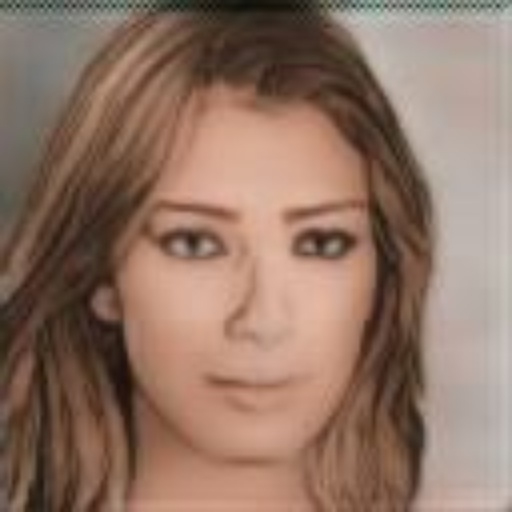}}
\end{minipage}%
\hspace{0.01\textwidth} 
\begin{minipage}[b]{0.14\textwidth}
  \centering
  \fbox{\includegraphics[width=\textwidth]{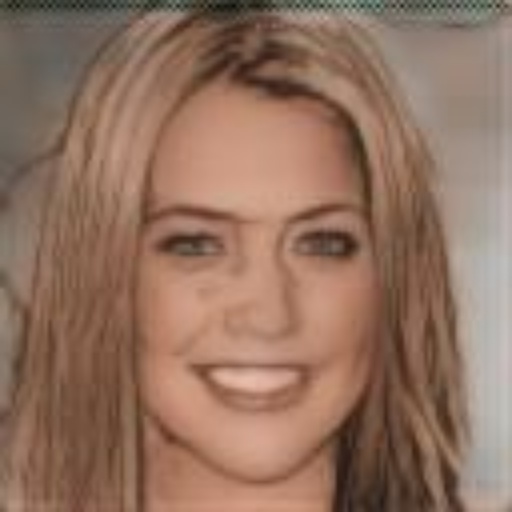}}
\end{minipage}%
\hspace{0.01\textwidth} 
\begin{minipage}[b]{0.14\textwidth}
  \centering
  \fbox{\includegraphics[width=\textwidth]{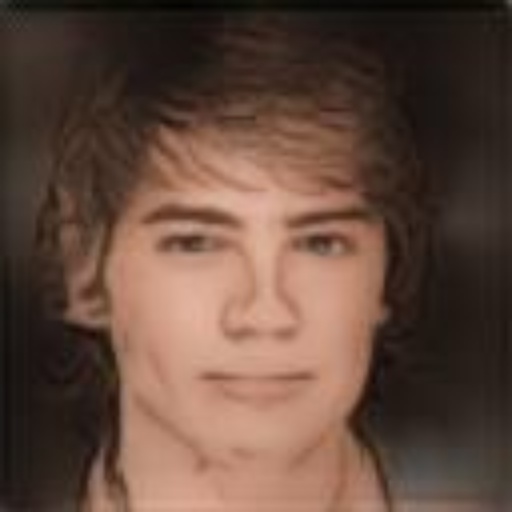}}
\end{minipage}%
\hspace{0.01\textwidth} 
\begin{minipage}[b]{0.14\textwidth}
  \centering
  \fbox{\includegraphics[width=\textwidth]{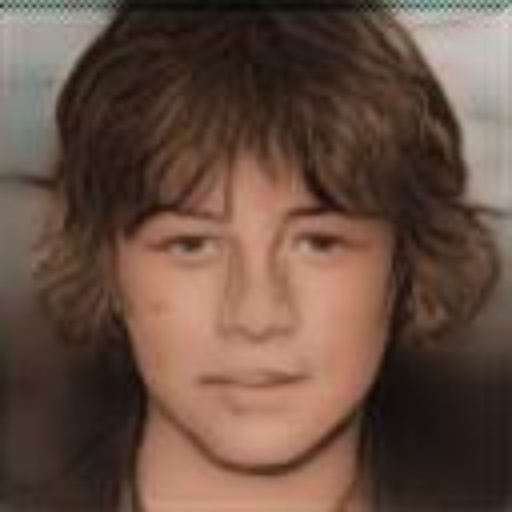}}
\end{minipage}%
\hspace{0.01\textwidth} 
\begin{minipage}[b]{0.14\textwidth}
  \centering
  \fbox{\includegraphics[width=\textwidth]{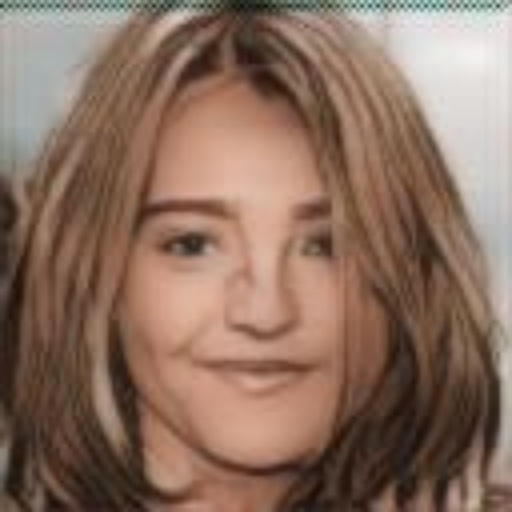}}
\end{minipage}%
\hspace{0.01\textwidth} 
\begin{minipage}[b]{0.14\textwidth}
  \centering
  \fbox{\includegraphics[width=\textwidth]{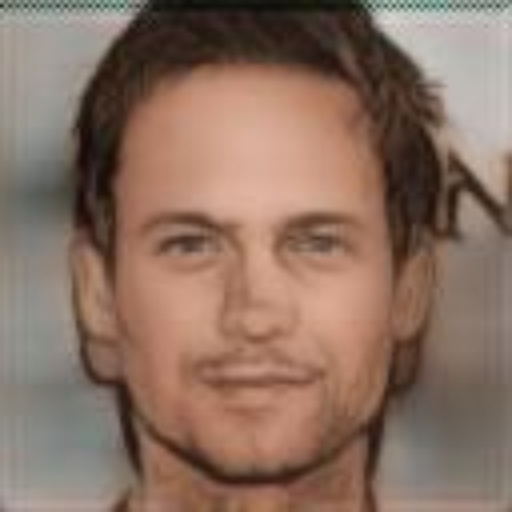}}
\end{minipage}

\vspace{0.5em} 

\rotatebox[origin=c]{90}{\parbox[c]{0\textheight}{\centering CycleGAN}}
\begin{minipage}[b]{0.14\textwidth}
  \centering
  \fbox{\includegraphics[width=\textwidth]{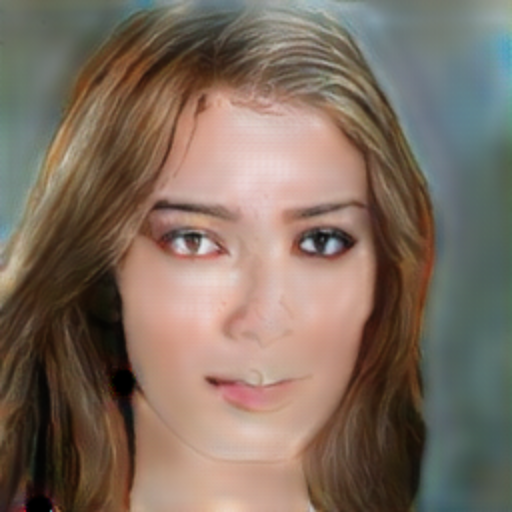}}
\end{minipage}%
\hspace{0.01\textwidth} 
\begin{minipage}[b]{0.14\textwidth}
  \centering
  \fbox{\includegraphics[width=\textwidth]{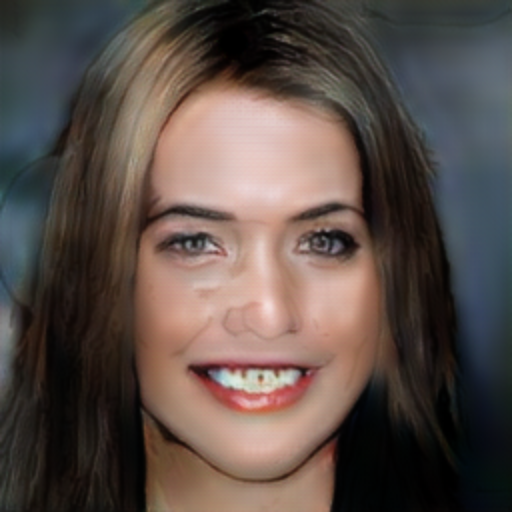}}
\end{minipage}%
\hspace{0.01\textwidth} 
\begin{minipage}[b]{0.14\textwidth}
  \centering
  \fbox{\includegraphics[width=\textwidth]{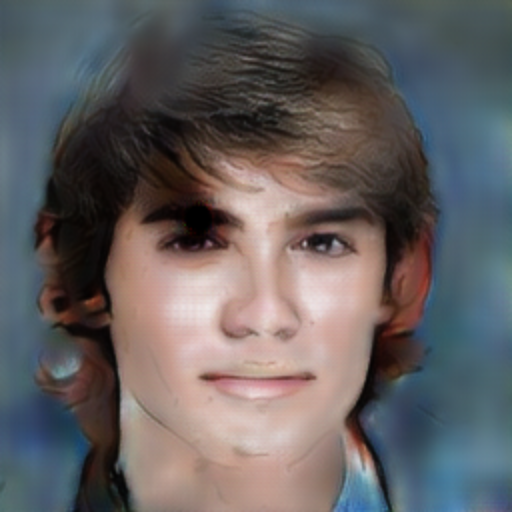}}
\end{minipage}%
\hspace{0.01\textwidth} 
\begin{minipage}[b]{0.14\textwidth}
  \centering
  \fbox{\includegraphics[width=\textwidth]{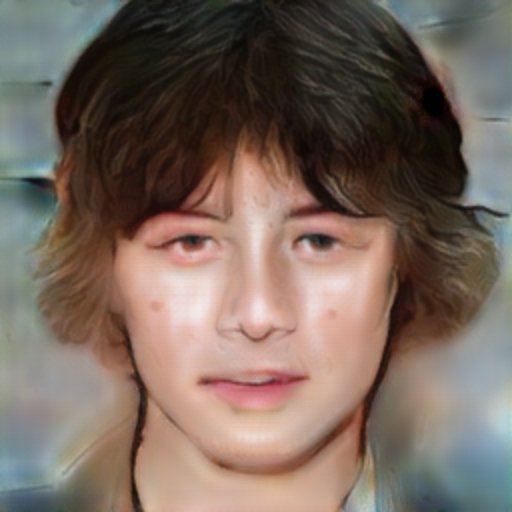}}
\end{minipage}%
\hspace{0.01\textwidth} 
\begin{minipage}[b]{0.14\textwidth}
  \centering
  \fbox{\includegraphics[width=\textwidth]{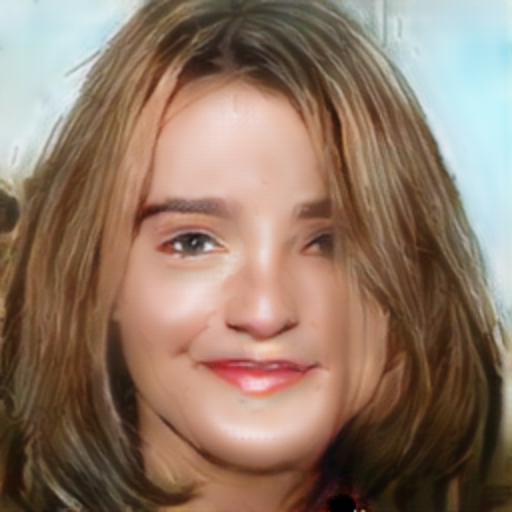}}
\end{minipage}%
\hspace{0.01\textwidth} 
\begin{minipage}[b]{0.14\textwidth}
  \centering
  \fbox{\includegraphics[width=\textwidth]{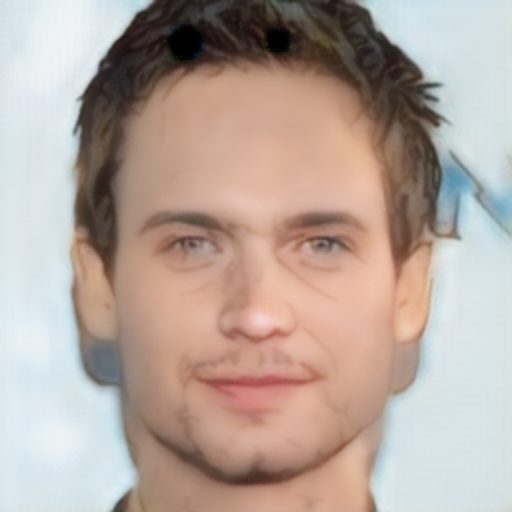}}
\end{minipage}

\vspace{0.5em} 

\rotatebox[origin=c]{90}{\parbox[c]{0\textheight}{\centering pSp}}
\begin{minipage}[b]{0.14\textwidth}
  \centering
  \fbox{\includegraphics[width=\textwidth]{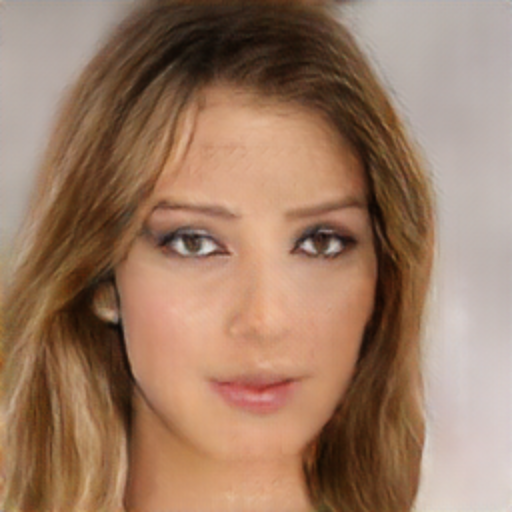}}
\end{minipage}%
\hspace{0.01\textwidth} 
\begin{minipage}[b]{0.14\textwidth}
  \centering
  \fbox{\includegraphics[width=\textwidth]{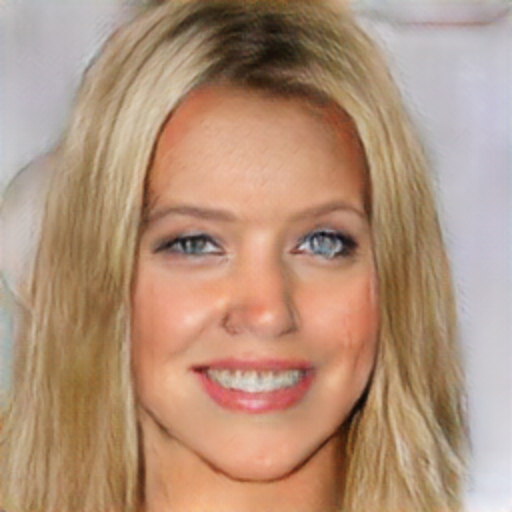}}
\end{minipage}%
\hspace{0.01\textwidth} 
\begin{minipage}[b]{0.14\textwidth}
  \centering
  \fbox{\includegraphics[width=\textwidth]{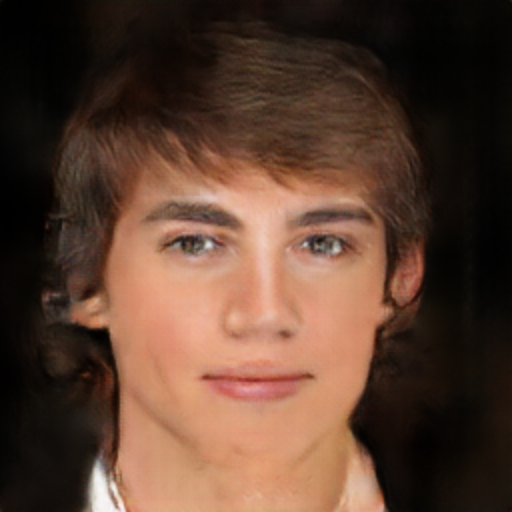}}
\end{minipage}%
\hspace{0.01\textwidth} 
\begin{minipage}[b]{0.14\textwidth}
  \centering
  \fbox{\includegraphics[width=\textwidth]{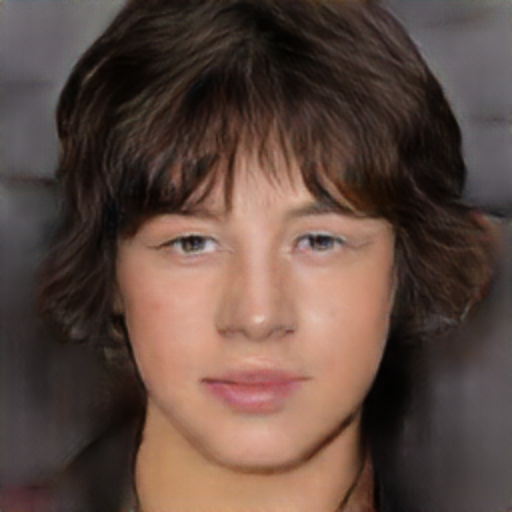}}
\end{minipage}%
\hspace{0.01\textwidth} 
\begin{minipage}[b]{0.14\textwidth}
  \centering
  \fbox{\includegraphics[width=\textwidth]{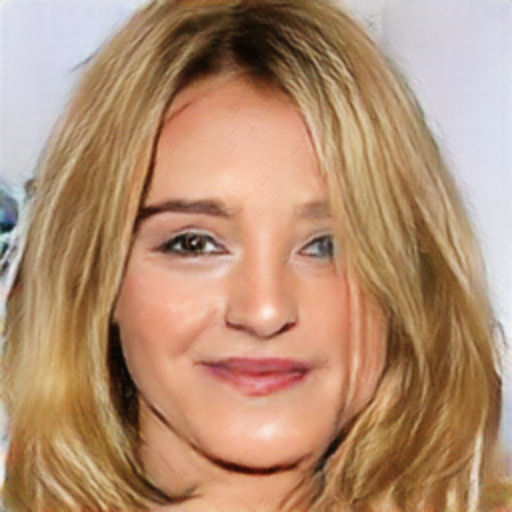}}
\end{minipage}%
\hspace{0.01\textwidth} 
\begin{minipage}[b]{0.14\textwidth}
  \centering
  \fbox{\includegraphics[width=\textwidth]{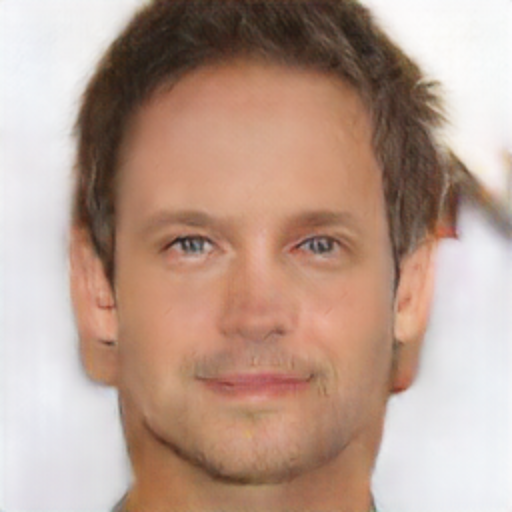}}
\end{minipage}

\vspace{1.5em} 

\rotatebox[origin=c]{90}{\parbox[c]{0\textheight}{\centering DFD}}
\begin{minipage}[b]{0.14\textwidth}
  \centering
  \fbox{\includegraphics[width=\textwidth]{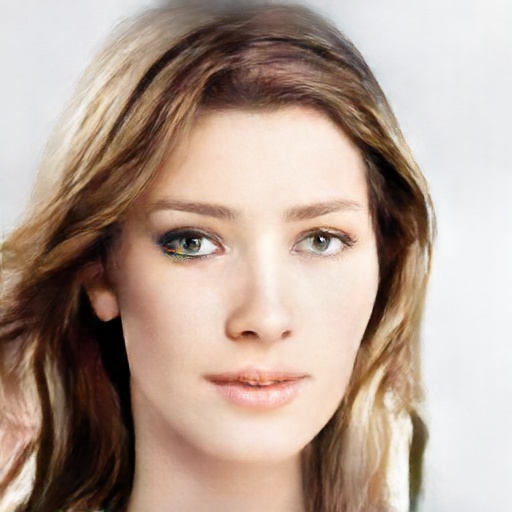}}
\end{minipage}%
\hspace{0.01\textwidth} 
\begin{minipage}[b]{0.14\textwidth}
  \centering
  \fbox{\includegraphics[width=\textwidth]{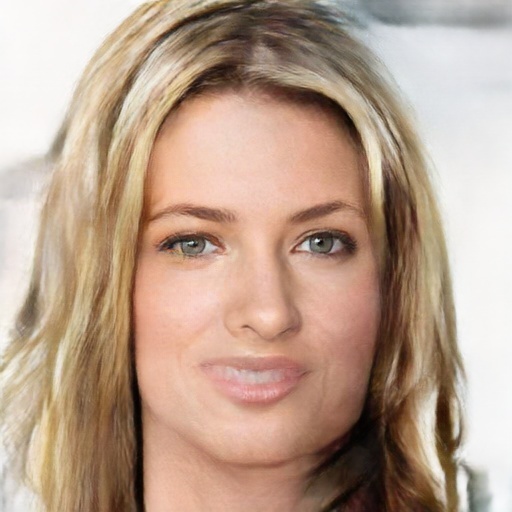}}
\end{minipage}%
\hspace{0.01\textwidth} 
\begin{minipage}[b]{0.14\textwidth}
  \centering
  \fbox{\includegraphics[width=\textwidth]{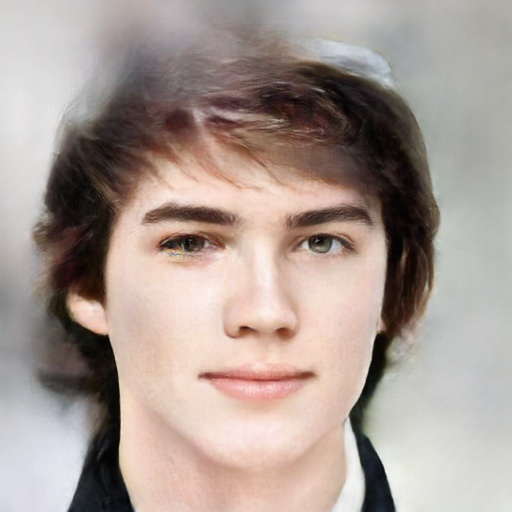}}
\end{minipage}%
\hspace{0.01\textwidth} 
\begin{minipage}[b]{0.14\textwidth}
  \centering
  \fbox{\includegraphics[width=\textwidth]{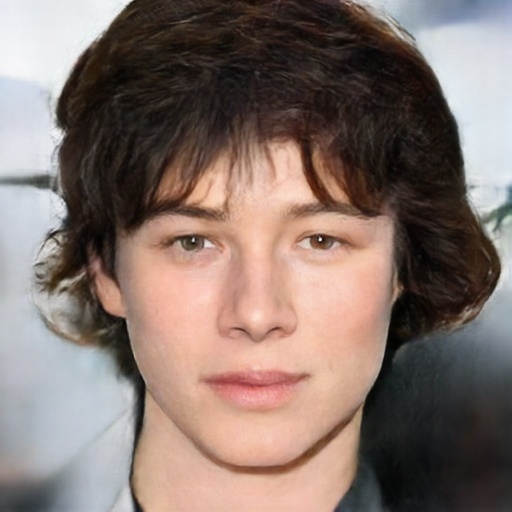}}
\end{minipage}%
\hspace{0.01\textwidth} 
\begin{minipage}[b]{0.14\textwidth}
  \centering
  \fbox{\includegraphics[width=\textwidth]{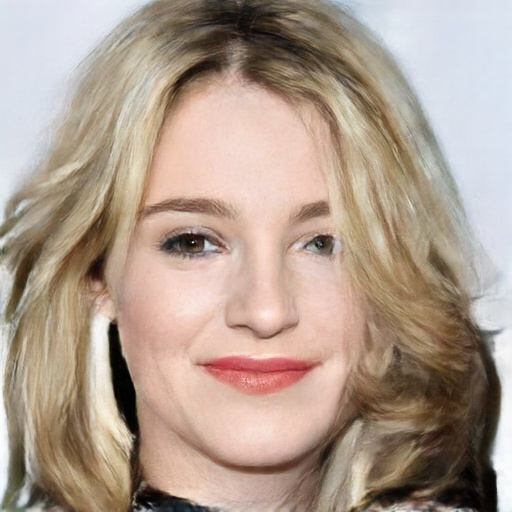}}
\end{minipage}%
\hspace{0.01\textwidth} 
\begin{minipage}[b]{0.14\textwidth}
  \centering
  \fbox{\includegraphics[width=\textwidth]{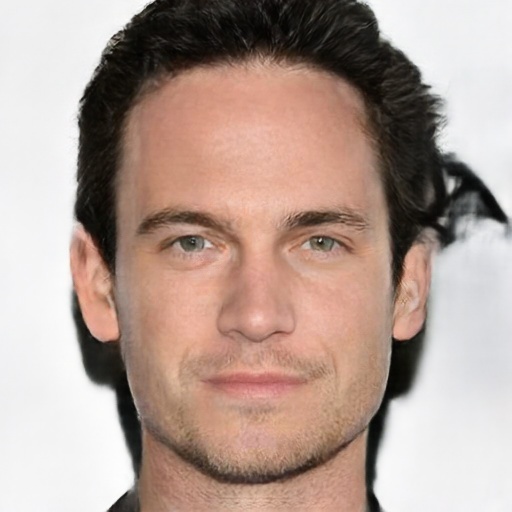}}
\end{minipage}

\vspace{0.5em} 

\rotatebox[origin=c]{90}{\parbox[c]{0\textheight}{\centering Ours}}
\begin{minipage}[b]{0.14\textwidth}
  \centering
  \fbox{\includegraphics[width=\textwidth]{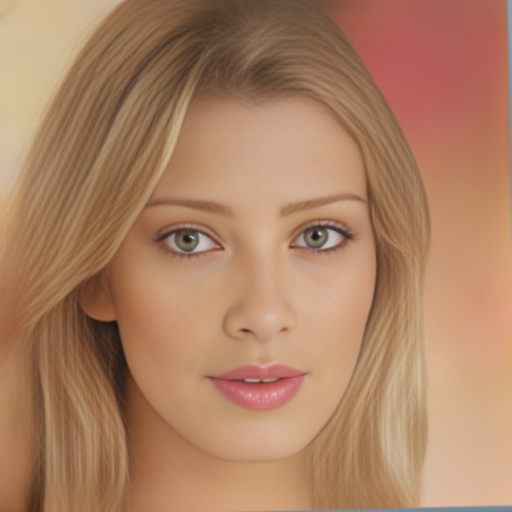}}
\end{minipage}%
\hspace{0.01\textwidth} 
\begin{minipage}[b]{0.14\textwidth}
  \centering
  \fbox{\includegraphics[width=\textwidth]{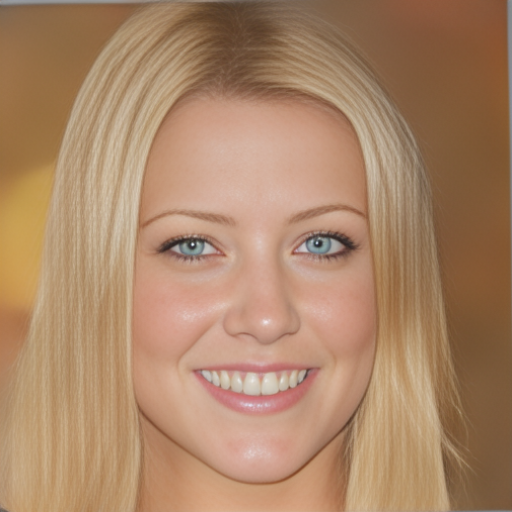}}
\end{minipage}%
\hspace{0.01\textwidth} 
\begin{minipage}[b]{0.14\textwidth}
  \centering
  \fbox{\includegraphics[width=\textwidth]{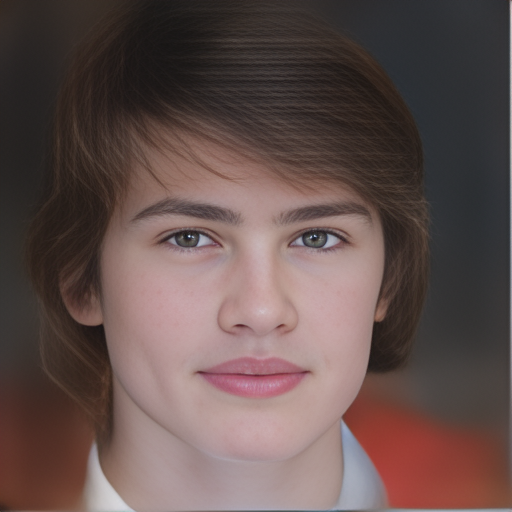}}
\end{minipage}%
\hspace{0.01\textwidth} 
\begin{minipage}[b]{0.14\textwidth}
  \centering
  \fbox{\includegraphics[width=\textwidth]{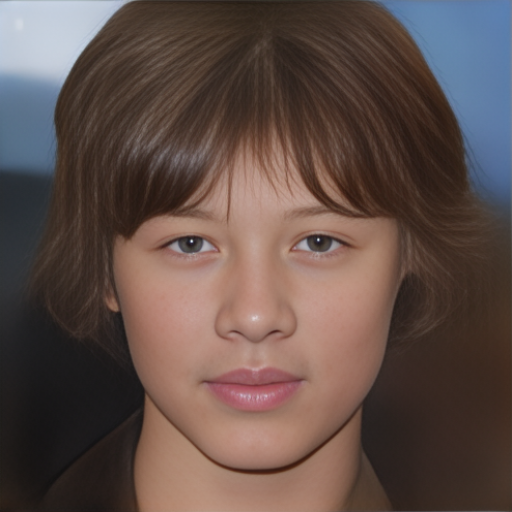}}
\end{minipage}%
\hspace{0.01\textwidth} 
\begin{minipage}[b]{0.14\textwidth}
  \centering
  \fbox{\includegraphics[width=\textwidth]{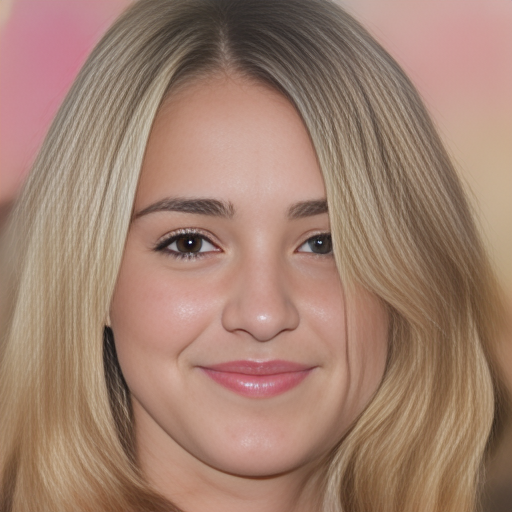}}
\end{minipage}%
\hspace{0.01\textwidth} 
\begin{minipage}[b]{0.14\textwidth}
  \centering
  \fbox{\includegraphics[width=\textwidth]{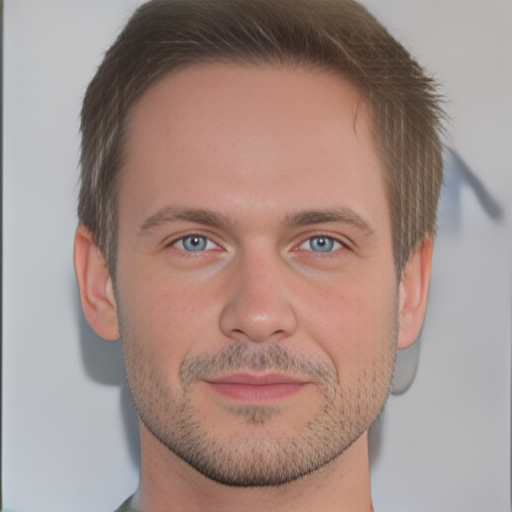}}
\end{minipage}

\caption{Qualitative comparison between our method and state-of-the-art approaches for sketch-to-image translation. Each column represents different sketch inputs and their corresponding outputs, generated by various methods.}
\label{fig:qualitative_results}
\end{figure*}
\subsubsection{Implementation Details}\label{subsubsec:implementation details}
In the proposed two-stage training process, each experiment is performed on a workstation having a NVIDIA GeForce RTX4080 GPU with 16GB of dedicated memory, 64GB RAM, and an Intel(R) Xeon(R) CPU E5-2667 v3 @ 3.20GHz. The first training stage requires 13.5GB of VRAM and 33 hours for 100 epochs with a batch size of 4. Furthermore, the second training stage requires 15.5GB of VRAM and 68 hours for 100 epochs with a batch size of 2. 

\subsubsection{Baseline Models and Rationale}\label{subsubsec:baseline}
We compare our method against a diverse set of baselines that represent both classical and recent advances in sketch-to-image translation and face generation. CycleGAN\cite{zhu2017unpaired} and Pix2PixHD\cite{wang2018high} are classical image-to-image translation methods, widely used as foundational baselines in this domain. pSp\cite{richardson2021encoding} and DeepFaceDrawing (DFD)\cite{chen2020deepfacedrawing} target semantic alignment and user-controllable synthesis, making them suitable for evaluating identity preservation and fine-grained reconstruction. OME\cite{guo2024image} introduces an optimisation-based multi-expert approach, improving generalisation across sketch variations.

More recently, ControlNet\cite{zhang2023adding} and T2I-Adapter\cite{mou2024t2i} extend diffusion-based generative models with conditional guidance, enabling state-of-the-art controllability and visual fidelity. Including these baselines allows us to benchmark our approach not only against traditional GAN-based frameworks but also against the latest diffusion-driven models, ensuring a fair and comprehensive comparison across paradigms.

\subsubsection{Evaluation Metrics}\label{subsubsec:evaluation metrics}

To assess the quality of our generated images, we used heuristic-based evaluation metrics that are widely used in sketch-to-image translation tasks. These metrics include Inception Score (IS) \cite{barratt2018note}, Fréchet Inception Distance (FID), Kernel Inception Distance (KID) \cite{betzalel2022study}, and Structural Similarity Index Measure (SSIM) \cite{zhang2018unreasonable}.

\begin{table}[t]
\centering
\caption{A comparison of the quantitative results of our approach with other baseline techniques across three different datasets. The best scores are highlighted in bold black, while the second-best scores are highlighted in bold blue.}
\label{tab:results}
\begin{tabular}{c@{\hspace{0.18cm}}c@{\hspace{0.18cm}}c@{\hspace{0.18cm}}c@{\hspace{0.18cm}}c@{\hspace{0.18cm}}c@{\hspace{0.18cm}}c} 
 \hline
 \textbf{Method} & \textbf{FID} $\downarrow$ & \textbf{IS} $\uparrow$  &  \textbf{KID} $\downarrow$ & \textbf{SSIM} $\uparrow$ &  \textbf{PSNR} $\uparrow$ & \textbf{LPIPS} $\downarrow$  \\ [0.5ex]
 \hline
 \hline
 \multicolumn{7}{c}{\textbf{CelebAMask-HQ dataset\cite{lee2020maskgan}}} \\
 \hline
ControlNet\cite{zhang2023adding} & 172.93 & \textbf{1.97} & 135.25 & 0.55 & 26.31 & 0.35 \\
T2I-Adapter\cite{mou2024t2i} & 508.00 & 1.00 & 114.76 & 0.31 & 16.27 & 0.48 \\
Pix2PixHD\cite{wang2018high} & 172.93 & 1.28 & 92.95 & 0.61 & 25.31 & 0.38 \\
CycleGAN\cite{zhu2017unpaired} & 148.78 & 1.31 & 105.99 & 0.53 & 28.02 & 0.28 \\
pSp\cite{richardson2021encoding} & 99.16 & 1.74 & 73.58 & \textbf{\textcolor{blue}{0.66}} & \textbf{29.32} & 0.17 \\
OME\cite{guo2024image} & - & - & - & 0.62 & 21.43 & \textbf{\textcolor{blue}{0.16}} \\
DFD\cite{chen2020deepfacedrawing} & \textbf{\textcolor{blue}{78.28}} & 1.25 & \textbf{\textcolor{blue}{61.28}} & 0.64 & 28.92 & 0.45 \\
\textbf{Ours} & \textbf{64.44} & \textbf{\textcolor{blue}{1.90}} & \textbf{43.26} & \textbf{0.77} & \textbf{\textcolor{blue}{28.99}} & \textbf{0.14} \\
 \hline
 \multicolumn{7}{c}{\textbf{CUHK dataset\cite{wang2008face}}} \\
 \hline
ControlNet\cite{zhang2023adding} & 254.43 & 1.00 & 109.10 & 0.51 & 23.87 & 0.36 \\
T2I-Adapter\cite{mou2024t2i} & 390.72 & 1.17 & 137.48 & 0.30 & 19.78 & 0.51 \\
Pix2PixHD\cite{wang2018high} & 157.71 & 1.34 & 72.50 & 0.68 & 26.77 & 0.25 \\
CycleGAN\cite{zhu2017unpaired} & 176.70 & 1.22 & 69.35 & 0.76 & 26.20 & 0.18 \\
pSp\cite{richardson2021encoding} & \textbf{\textcolor{blue}{129.41}} & \textbf{\textcolor{blue}{1.37}} & \textbf{34.12} & \textbf{\textcolor{blue}{0.78}} & \textbf{\textcolor{blue}{27.61}} & \textbf{\textcolor{blue}{0.17}} \\
\textbf{Ours} & \textbf{84.68} & \textbf{1.39} & \textbf{\textcolor{blue}{35.06}} & \textbf{0.79} & \textbf{28.95} & \textbf{0.18} \\
\hline
 \multicolumn{7}{c}{\textbf{CUFSF dataset\cite{zhang2011coupled}}} \\
 \hline
ControlNet\cite{zhang2023adding} & 256.71 & 1.20 & 109.01 & 0.42 & 23.80 & 0.44 \\
T2I-Adapter\cite{mou2024t2i} & 195.87 & 1.08 & 102.85 & 0.52 & 25.62 & 0.37 \\
Pix2PixHD\cite{wang2018high} & 172.96 & \textbf{1.68} & 128.08 & 0.66 & 26.31 & 0.26 \\
CycleGAN\cite{zhu2017unpaired} & \textbf{\textcolor{blue}{148.75}} & 1.41 & \textbf{\textcolor{blue}{78.24}} & 0.69 & \textbf{\textcolor{blue}{27.03}} & 0.24 \\
pSp\cite{richardson2021encoding} & 159.69 & \textbf{\textcolor{blue}{1.67}} & 96.63 & \textbf{0.77} & 26.71 & \textbf{0.17} \\
\textbf{Ours} & \textbf{78.48} & 1.62 & \textbf{76.49} & \textbf{\textcolor{blue}{0.76}} & \textbf{29.14} & \textbf{\textcolor{blue}{0.20}} \\
 \hline
\end{tabular}
\end{table}

\subsection{Quantitative Results}\label{subsec:quantitative results}
To quantify the quality of our generated images, we used heuristic-based evaluation measures such as Inception Score (IS)~\cite{barratt2018note}, Fréchet Inception Distance (FID), and Kernel Inception Distance (KID)~\cite{betzalel2022study} and Structural Similarity Index Measure (SSIM)~\cite{hore2010image} to asses the structural similarity between generated and original images.
To evaluate the quality of our generated images, we employed heuristic-based evaluation measures commonly used in image-to-image translation tasks, including Inception Score (IS), Fréchet Inception Distance (FID), Kernel Inception Distance (KID), Peak Signal-to-Noise Ratio (PSNR)~\cite{mohammadi2014subjective}, Learned Perceptual Image Patch Similarity(LPIPS)~\cite{zhang2018unreasonable} and Structural Similarity Index Measure (SSIM) \cite{betzalel2022study}. These metrics provide insights into various aspects of the generated images, such as diversity, fidelity, and structural similarity. We first provide a quantitative comparison against state-of-the-art techniques and our framework~\cite{chen2020deepfacedrawing} in Table \ref{tab:results}.

We present a quantitative comparison of our framework with several state-of-the-art GAN-based and diffusion-based methods on three benchmark datasets, as detailed in Table \ref{tab:results}. 
The techniques compared include ControlNet \cite{zhang2023adding}, T2I-Adapter \cite{mou2024t2i}, Pix2PixHD \cite{wang2018high}, CycleGAN \cite{zhu2017unpaired}, pSp \cite{richardson2021encoding}, OME \cite{guo2024image}, and DeepFaceDrawing (DFD) \cite{chen2020deepfacedrawing}. Our framework demonstrates superior performance in all metrics evaluated, achieving significantly lower FID and KID scores, along with higher IS and SSIM values. These results highlight the effectiveness of our approach in generating faces that closely resemble real ones in both appearance and structural similarity.

On the CelebAMask-HQ dataset, our framework achieved improvements of 21\%, 58\%, 41\%, and 20\% in FID, IS, KID, and SSIM scores, respectively, when compared to DFD. Furthermore, our method outperforms other state-of-the-art approaches, as detailed in Table \ref{tab:results}. We also trained our framework on the CUHK and CUFSF datasets and benchmarked its performance against techniques such as pSp, CycleGAN, and Pix2PixHD. Due to the unavailability of training codes for the OME and DFD methods, we could only utilise their pre-trained weights for the CelebAMask-HQ dataset.

In the CUHK dataset, our framework demonstrated a 53\% improvement in FID compared to pSp. However, pSp slightly outperformed our method in terms of the KID score. In the CUFSF data set, CycleGAN is the second-best algorithm in terms of FID and KID scores. Our framework improved FID by 89\% and KID by 2.6\% relative to the CycleGAN method. However, PIx2PIxHD exhibited better performance in the CUFSF data set in terms of IS, as illustrated in Table \ref{tab:results}.

\subsection{Qualitative Results}
Our qualitative analysis further validates the quantitative results, as depicted in Figure \ref{fig:qualitative_results}. The images generated by Pix2PixHD and pSp correspond to input sketches, which means these methods generate the local information effectively. However, the generated images are of low quality, and Pix2PixHD missed the colour information. In contrast, the CycleGAN generates the colour information, however, if the images have a background colour other than white, then the CycleGAN mixes the background colour information with the foreground face image at the boundary, as shown in (c) image of CycleGAN in Figure \ref{fig:qualitative_results}. Furthermore, the images generated by DeepFaceDrawing are relatively bright, and these images do not correspond to ground truth labels as shown in (b) of DFD in Figure \ref{fig:qualitative_results}. Moreover, (e) DeepFaceDrawing generates different hairstyles. The results generated by our framework cover the limitations of previous state-of-the-art models. The image generated by our model captures the colour information, and the images are of high quality. Most importantly, these images correspond to the input sketch, as shown in Figure \ref{fig:qualitative_results}. Specifically, image (b) of our method captures the facial expressions very well, and the jawline of the generated image is very sharp, which is missing in other methods. Moreover, in (e) of our method, in generated the hairstyle that corresponds to the input sketch. However, in (d) image in Figure \ref{fig:qualitative_results}, the person has moles on his face, and all the methods, including ours unable to capture this mole. This may happen due to the limited information in the sketch, and the model treats these moles as noise. The results show that our framework generates faces with a higher degree of realism compared to both DeepFaceDrawing and other sketch-to-image synthesis techniques.

\begin{figure}[http]
\centering

\rotatebox[origin=c]{90}{\parbox[c]{0\textheight}{\centering Input}}
\begin{minipage}[b]{0.14\textwidth}
  \centering
  \makebox[0pt]{Line Sketch}\\
  \fbox{\includegraphics[width=\textwidth]{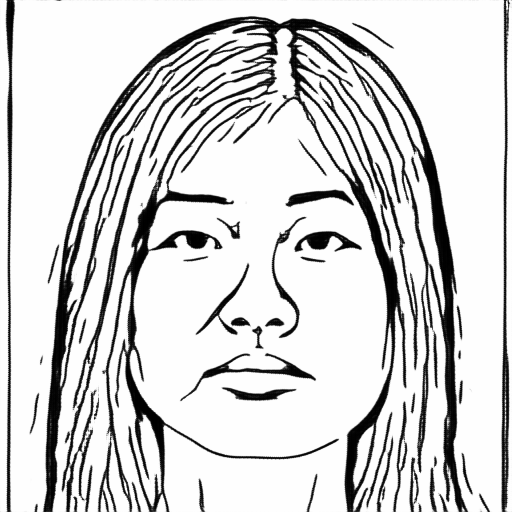}}
\end{minipage}%
\hspace{0.01\textwidth} 
\begin{minipage}[b]{0.14\textwidth}
  \centering
  \makebox[0pt]{Photoshop Sketch}\\
  \fbox{\includegraphics[width=\textwidth]{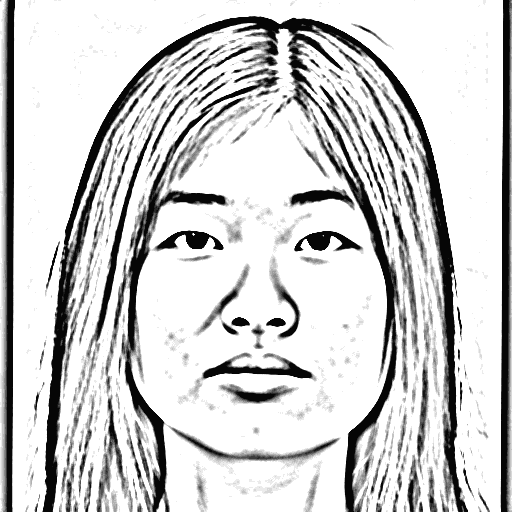}}
\end{minipage}%
\hspace{0.01\textwidth} 
\begin{minipage}[b]{0.14\textwidth}
  \centering
  \makebox[0pt]{ Hand-drawn Sketch}\\
  \fbox{\includegraphics[width=\textwidth]{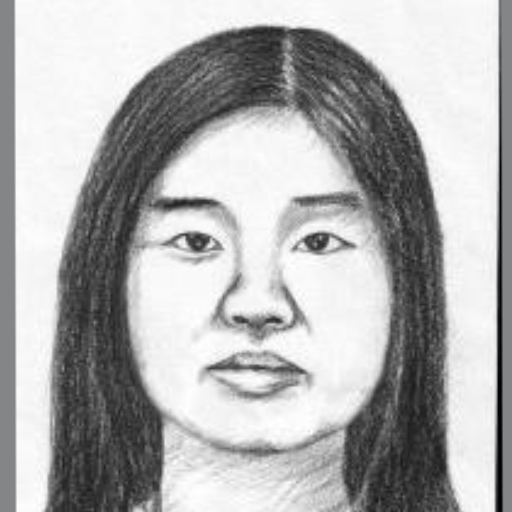}}
\end{minipage}%

\rotatebox[origin=c]{90}{\parbox[c]{0\textheight}{\centering GT}}
\begin{minipage}[b]{0.14\textwidth}
  \centering
  \fbox{\includegraphics[width=\textwidth]{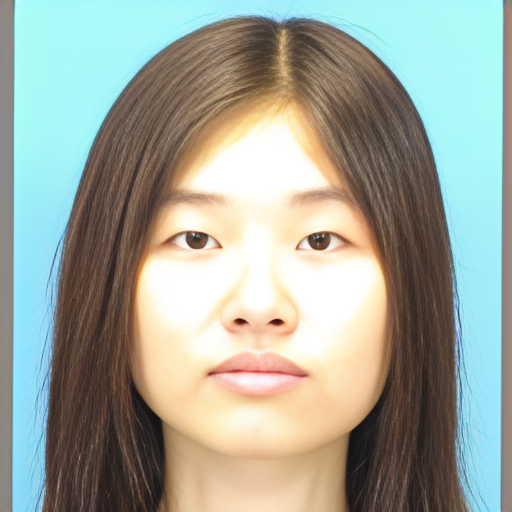}}
\end{minipage}%
\hspace{0.01\textwidth} 
\begin{minipage}[b]{0.14\textwidth}
  \centering
  \fbox{\includegraphics[width=\textwidth]{different_sketches/Different_S/GT.png}}
\end{minipage}%
\hspace{0.01\textwidth} 
\begin{minipage}[b]{0.14\textwidth}
  \centering
  \fbox{\includegraphics[width=\textwidth]{different_sketches/Different_S/GT.png}}
\end{minipage}%

\rotatebox[origin=c]{90}{\parbox[c]{0\textheight}{\centering DFD}}
\begin{minipage}[b]{0.14\textwidth}
  \centering
  \fbox{\includegraphics[width=\textwidth]{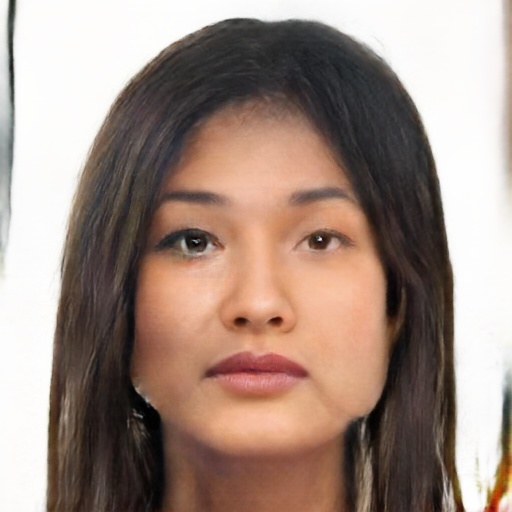}}
\end{minipage}%
\hspace{0.01\textwidth} 
\begin{minipage}[b]{0.14\textwidth}
  \centering
  \fbox{\includegraphics[width=\textwidth]{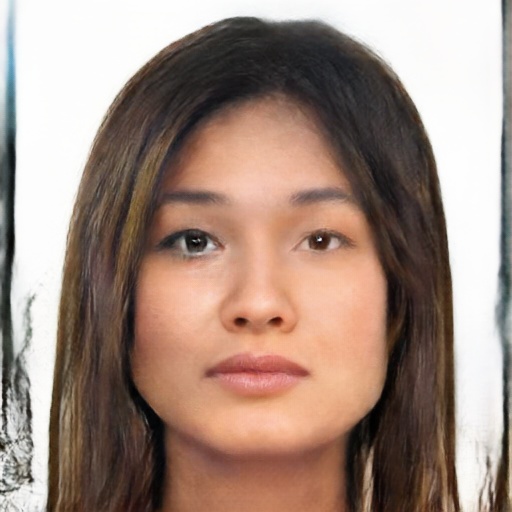}}
\end{minipage}%
\hspace{0.01\textwidth} 
\begin{minipage}[b]{0.14\textwidth}
  \centering
  \fbox{\includegraphics[width=\textwidth]{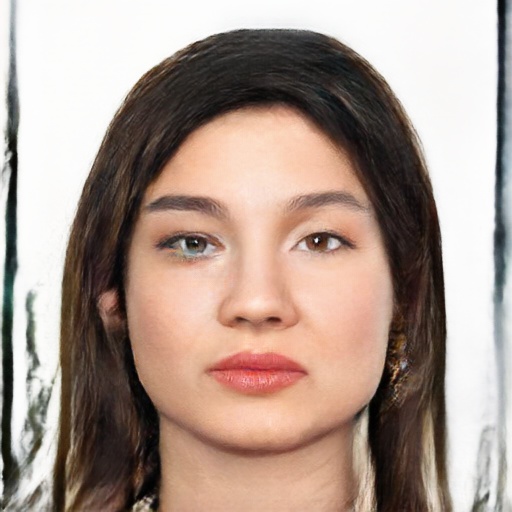}}
\end{minipage}%

\rotatebox[origin=c]{90}{\parbox[c]{0\textheight}{\centering Ours}}
\begin{minipage}[b]{0.14\textwidth}
  \centering
  \fbox{\includegraphics[width=\textwidth]{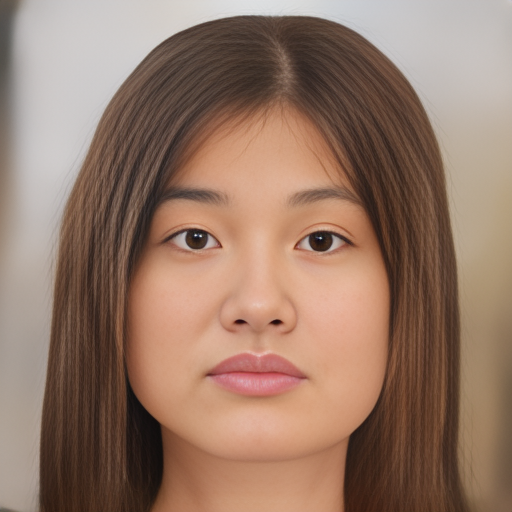}}
\end{minipage}%
\hspace{0.01\textwidth} 
\begin{minipage}[b]{0.14\textwidth}
  \centering
  \fbox{\includegraphics[width=\textwidth]{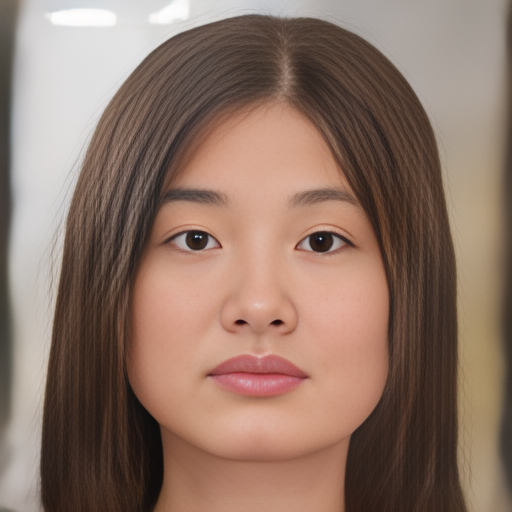}}
\end{minipage}%
\hspace{0.01\textwidth} 
\begin{minipage}[b]{0.14\textwidth}
  \centering
  \fbox{\includegraphics[width=\textwidth]{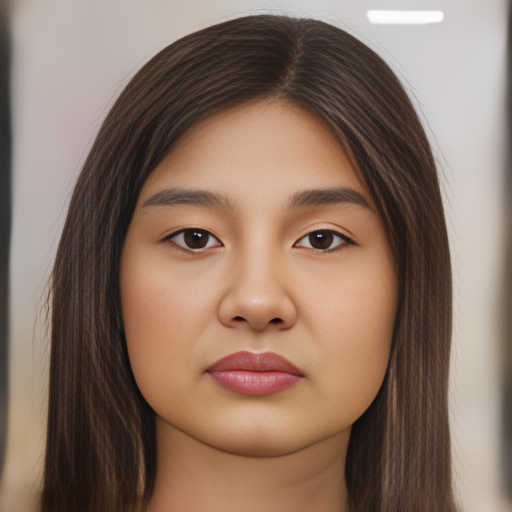}}
\end{minipage}%

\caption{Zero-shot comparison of DFD and Ground Truth (GT) with our method, trained on the CelebAMask-HQ dataset and tested on diverse sketch types.}
\label{fig:different sketches results}
\end{figure}

Our framework demonstrates strong generalisation for various types of sketches, including line sketches, Photoshop-generated sketches, and hand-drawn sketches, as illustrated in Figure \ref{fig:different sketches results}. The generated images closely match the original sketches, particularly for line and Photoshop sketches, which tend to be more precise compared to hand-drawn sketches. Our framework generates faces with a higher degree of realism compared to both DFD and other sketch-to-image synthesis techniques. The quantitative results in Table \ref{tab:domain_adaptation_results} and the qualitative results in Figure \ref{fig:different sketches results} demonstrate the superior generalisation ability of our model compared to DFD.


\begin{table}[t]
\centering
\caption{The comparison of our framework with DFD across different sketch types highlights the domain invariance of our approach.}
\label{tab:domain_adaptation_results}
\begin{tabular}{c@{\hspace{0.18cm}}c@{\hspace{0.18cm}}c@{\hspace{0.18cm}}c@{\hspace{0.18cm}}c@{\hspace{0.18cm}}c@{\hspace{0.18cm}}c@{\hspace{0.18cm}}c} 

 \hline
 \textbf{Sketch Type} & \textbf{FID} $\downarrow$ & \textbf{IS} $\uparrow$  &  \textbf{KID} $\downarrow$ & \textbf{SSIM} $\uparrow$ & \textbf{PSNR} $\uparrow$ & \textbf{LPIPS} $\downarrow$ \\ [0.5ex]
 \hline
 \hline
 \multicolumn{6}{c}{\textbf{DFD\cite{chen2020deepfacedrawing}}} \\
   \hline
 Hand Drawn Sketch & 191.01 & 1.70    & 94.65  & 0.52 & 25.768 & 0.368 \\
 Line Sketch & 130.88 & 1.49    & 38.87  & 0.67 & 27.570 & 0.259  \\
 Photoshop Sketch & 115.63 & 1.51  & 59.15 & 0.65 & 28.027 & 0.278 \\

 \hline
  \multicolumn{6}{c}{\textbf{Ours}} \\
 \hline
  Hand Drawn Sketch & 131.26 & 1.63  & 63.12  & 0.67 & 27.558 & 0.259 \\

Line Sketch  & 88.64 & 1.59    & 41.56 & 0.76  & 28.836 & 0.195 \\

Photoshop Sketch & 82.97 &  1.69 & 44.78  & 0.74  & 29.006 & 0.212 \\

 \hline

\end{tabular}
\end{table}

\subsection{Results on Different type of Sketches}\label{subsubsec:different sketch results}
To demonstrate our framework's generalisation capabilities, we present results across three distinct types of sketches, each differing in style, complexity, and origin. These types include hand-drawn sketches, line sketches, and Photoshop-generated sketches. Human artists produce hand-drawn sketches, whereas line and Photoshop sketches are generated through computer-aided methods.

The results of our framework, trained on the CelebAMask-HQ dataset and tested on 188 hand-drawn sketches from the CUHK dataset, are presented in Table \ref{tab:domain_adaptation_results}. In addition, the other two types of sketch datasets were generated using 188 images from the CUHK dataset. The performance of the DFD method on these sketch types is also provided in Table \ref{tab:domain_adaptation_results} for comparison.

Our framework exhibited robust performance across all three sketch categories and evaluation metrics. Specifically, in the Photoshop and line sketch categories, our method achieved an improvement of 40\% and 47\%, respectively, in the FID score compared to DFD. This substantial gain is due to the higher fidelity of computer-aided sketches, which more closely mirror the original images. In contrast, hand-drawn sketches produced by artists present more variability and imperfections. These findings illustrate the effectiveness of our framework in generalising across different types of sketches and its capability to generate high-quality images.

\subsection{Results on Non-Facial Datasets}\label{subsubsec:Non-Facial sketch results}
Our framework is primarily designed for facial datasets. However, for non-facial datasets, we modified the first training stage to better capture structural details. Specifically, we adopt the Tactile Sketch Saliency (TSS) model \cite{jiao2020tactile} as an auxiliary prior to enhance structural awareness in sketch representations. TSS generates saliency maps that highlight perceptually important regions. However, its continuous outputs often yield diffuse activations that lack sharp object boundaries. To address this, we discretise the saliency responses through multi-level thresholding and subsequently apply clustering to consolidate salient regions into compact, boundary preserving groups. We employ DBSCAN, as it adaptively captures irregular, non-convex contours and suppresses noisy activations without requiring a predefined number of clusters. This refinement enhances contour fidelity, reduces background noise, and provides sharper structural cues, enabling the autoencoder to better preserve object boundaries and improve the quality of learned sketch features.

\begin{figure}[htbp]
    \centering

    \begin{tabular}{c|ccc}
        \hline
        \multicolumn{4}{c}{\textbf{Sketchy Database~\cite{sangkloy2016sketchy}}} \\
        \hline
        \rotatebox{90}{Sketches} &
        \includegraphics[width=0.18\linewidth]{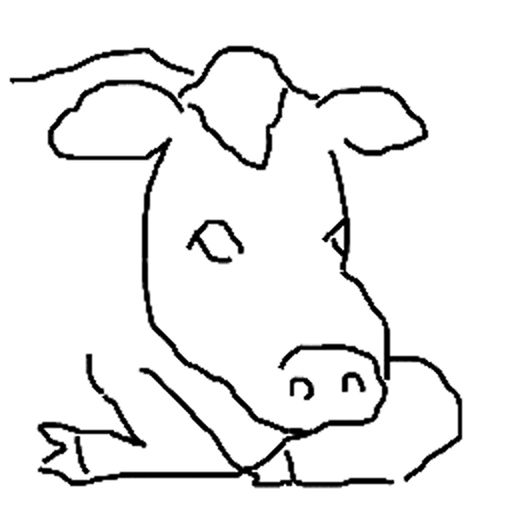} &
        \includegraphics[width=0.18\linewidth]{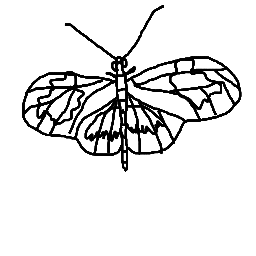} &
        \includegraphics[width=0.18\linewidth]{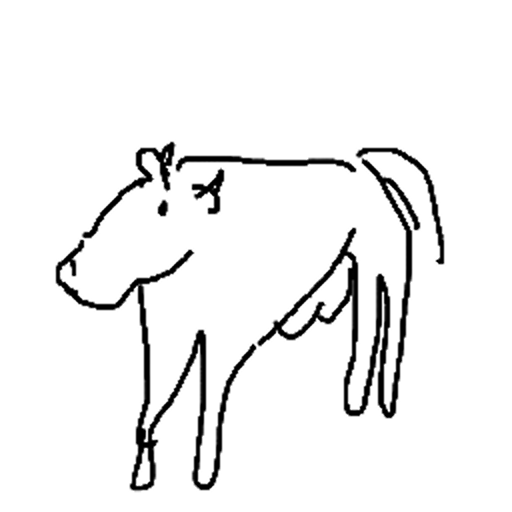} \\
        \rotatebox{90}{ GT} &
        \includegraphics[width=0.18\linewidth]{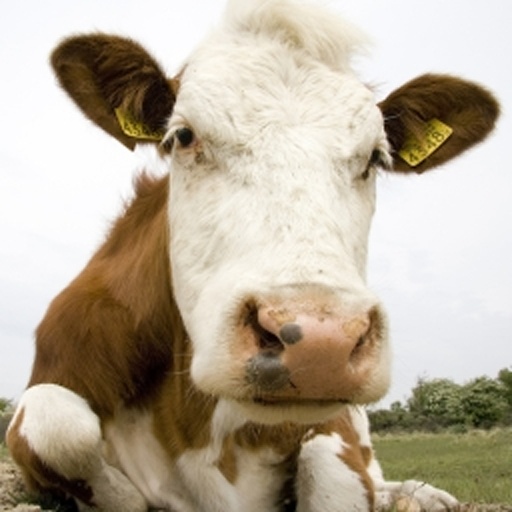} &
        \includegraphics[width=0.18\linewidth]{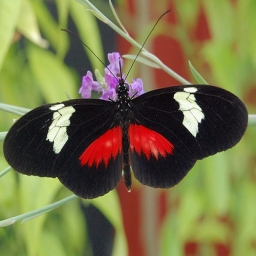} &
        \includegraphics[width=0.18\linewidth]{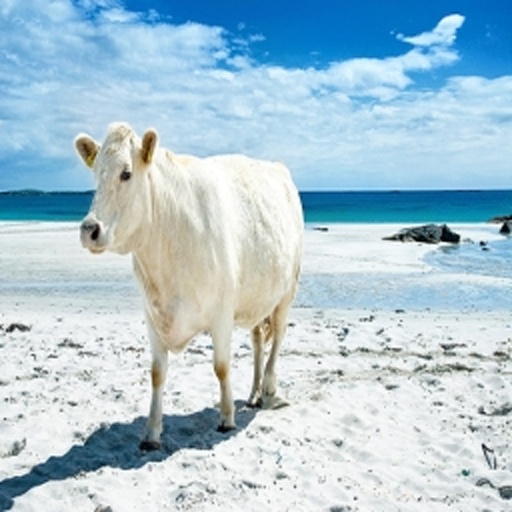} \\
        \rotatebox{90}{CycleGAN} &
        \includegraphics[width=0.18\linewidth]{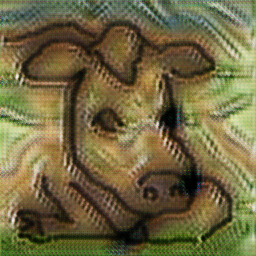} &
        \includegraphics[width=0.18\linewidth]{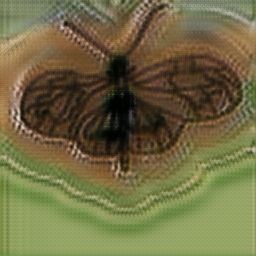} &
        \includegraphics[width=0.18\linewidth]{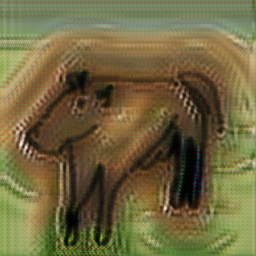} \\
        \rotatebox{90}{Ours} &
        \includegraphics[width=0.18\linewidth]{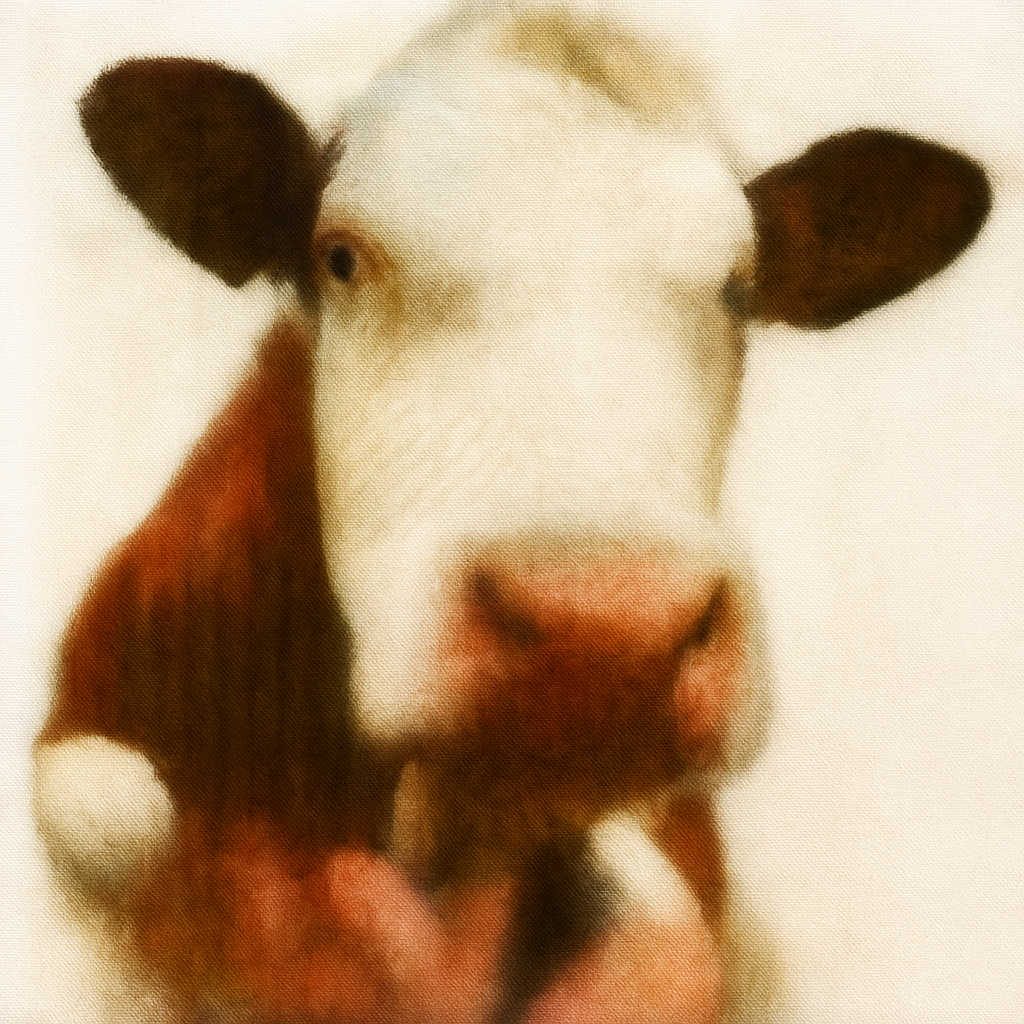} &
        \includegraphics[width=0.18\linewidth]{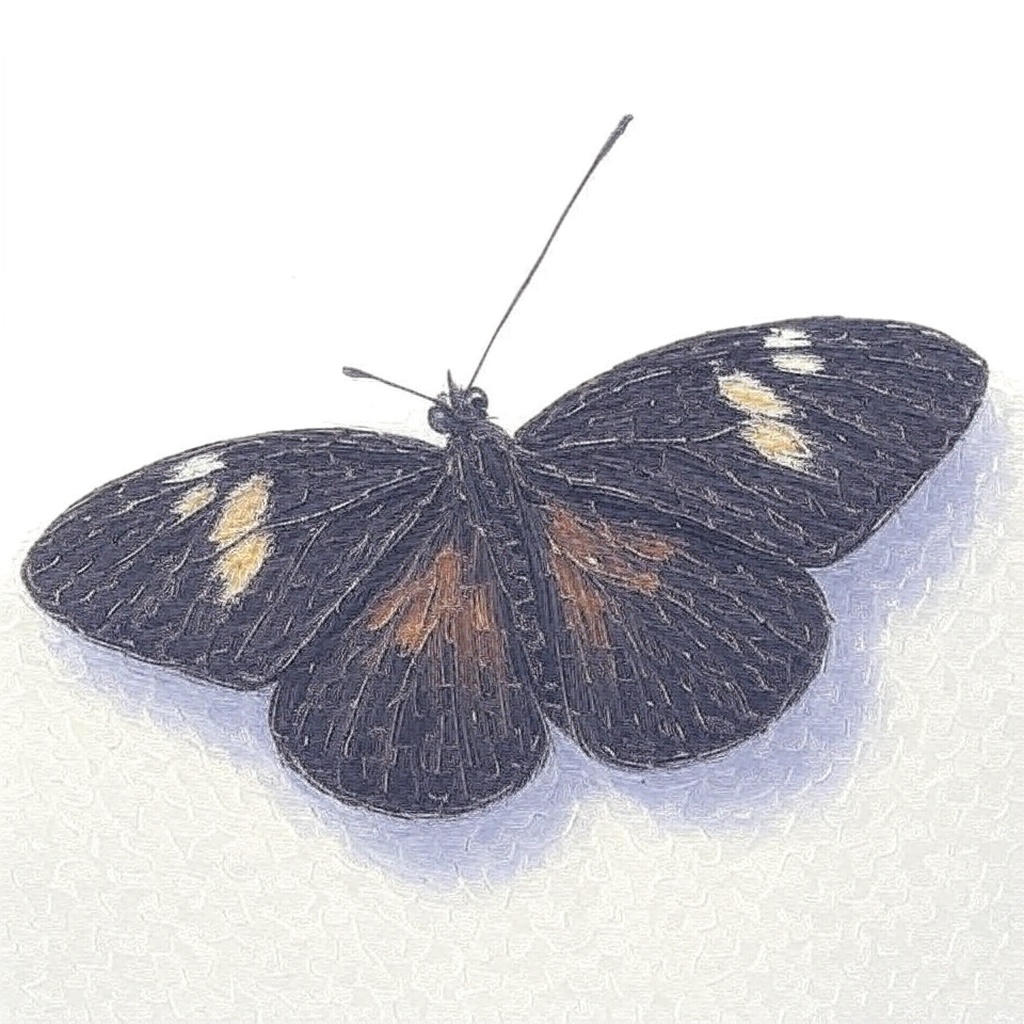} &
        \includegraphics[width=0.18\linewidth]{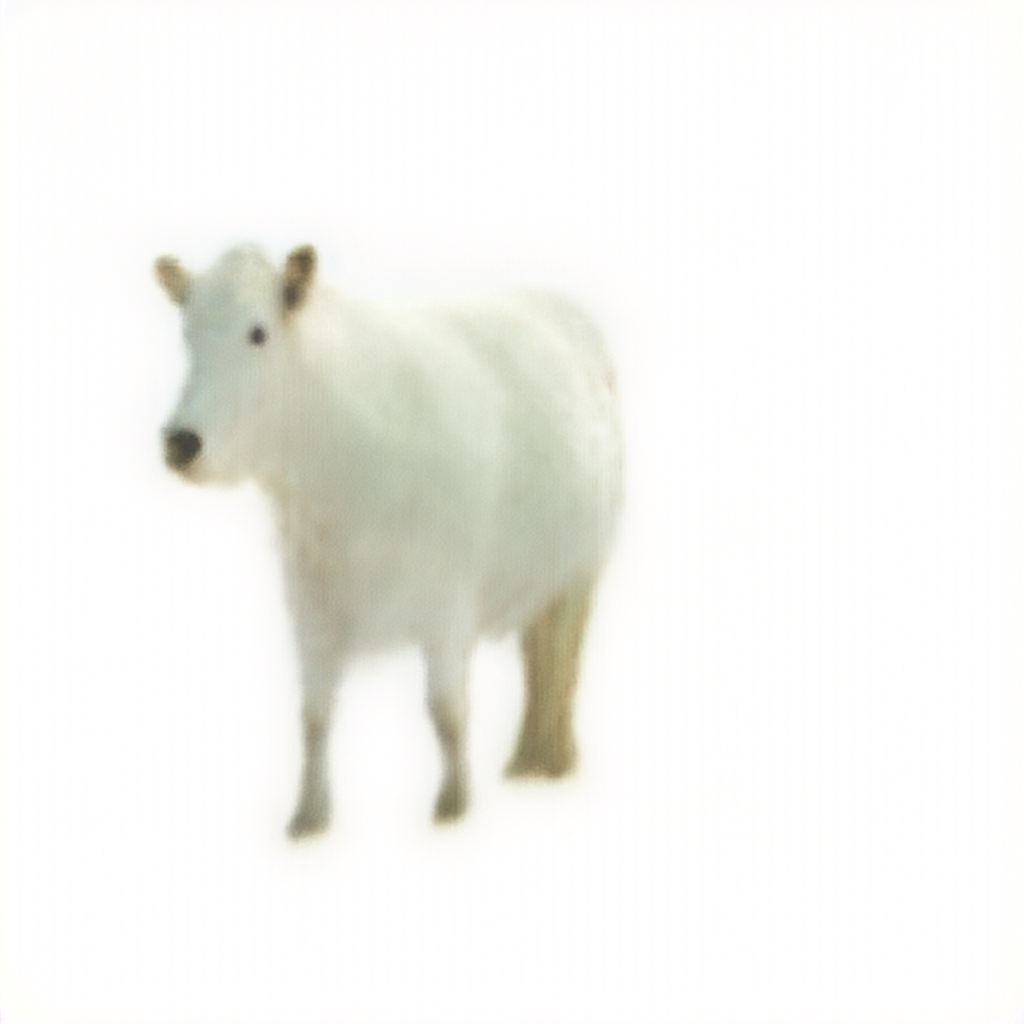} \\
    \end{tabular}
    \vspace{1em}

    \begin{tabular}{c|ccc}
        \hline
        \multicolumn{4}{c}{\textbf{QMUL ShoesV2~\cite{yu2021fine,bhunia2020sketch}}} \\
        \hline
        \rotatebox{90}{Sketches} &
        \includegraphics[width=0.18\linewidth]{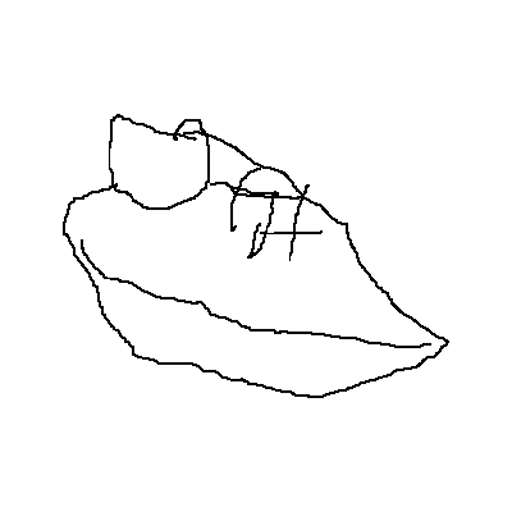} &
        \includegraphics[width=0.18\linewidth]{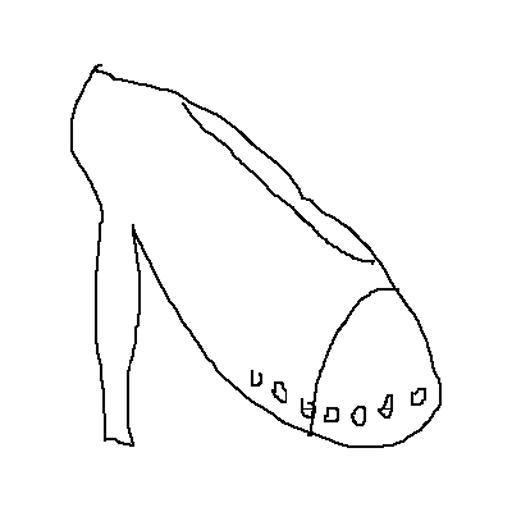} &
        \includegraphics[width=0.18\linewidth]{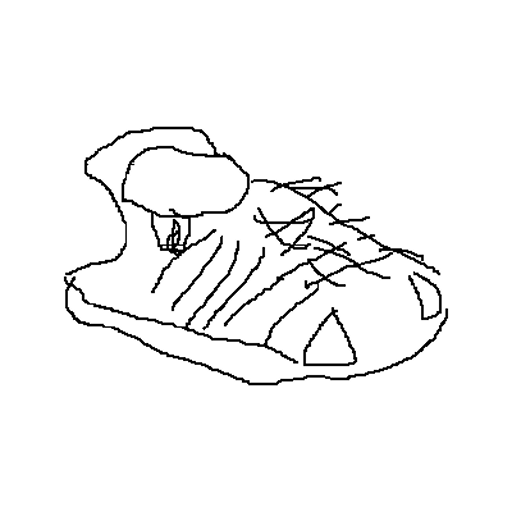} \\
        \rotatebox{90}{ GT} &
        \includegraphics[width=0.18\linewidth]{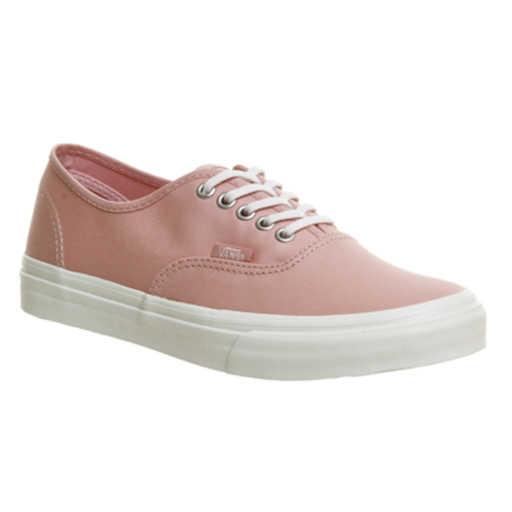} &
        \includegraphics[width=0.18\linewidth]{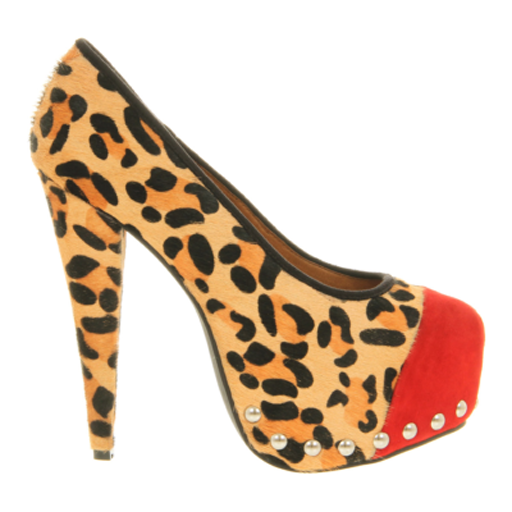} &
        \includegraphics[width=0.18\linewidth]{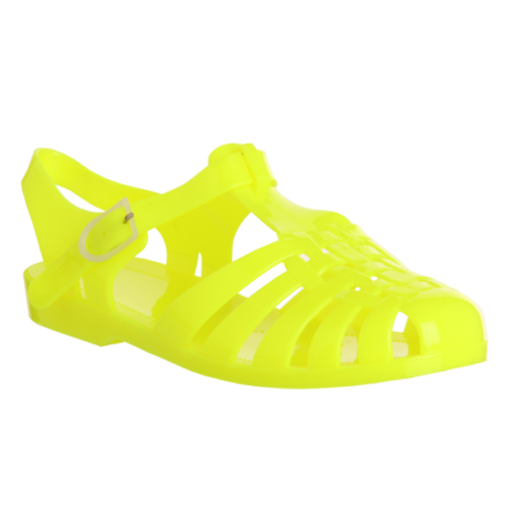} \\
        \rotatebox{90}{CycleGAN} &
        \includegraphics[width=0.18\linewidth]{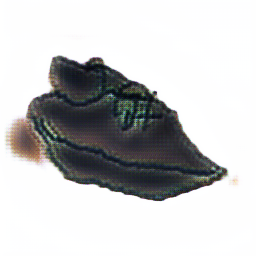} &
        \includegraphics[width=0.18\linewidth]{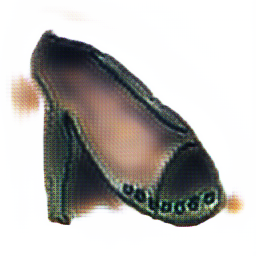} &
        \includegraphics[width=0.18\linewidth]{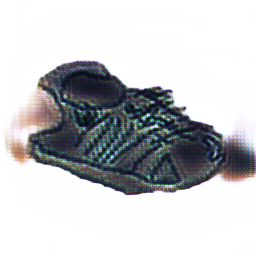} \\
        \rotatebox{90}{Ours} &
        \includegraphics[width=0.18\linewidth]{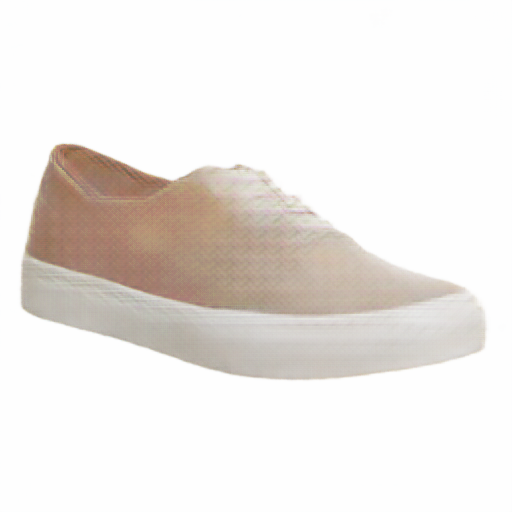} &
        \includegraphics[width=0.18\linewidth]{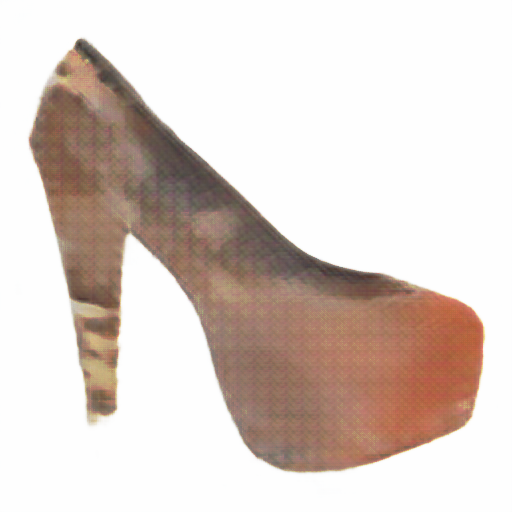} &
        \includegraphics[width=0.18\linewidth]{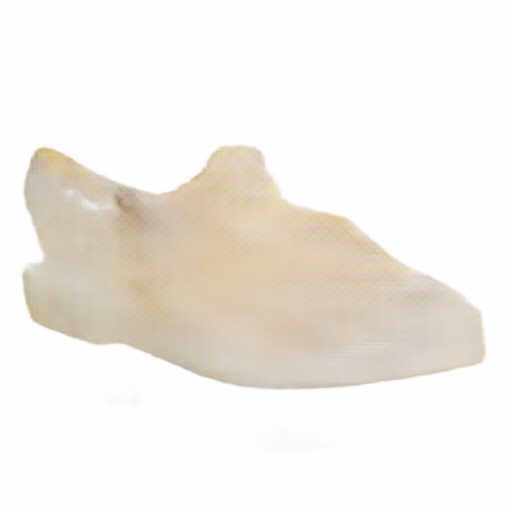} \\
    \end{tabular}
    \vspace{1em}

    \begin{tabular}{c|ccc}
        \hline
        \multicolumn{4}{c}{\textbf{QMUL ChairsV2~\cite{yu2021fine,bhunia2020sketch}}} \\
        \hline
        \rotatebox{90}{Sketches} &
        \includegraphics[width=0.18\linewidth]{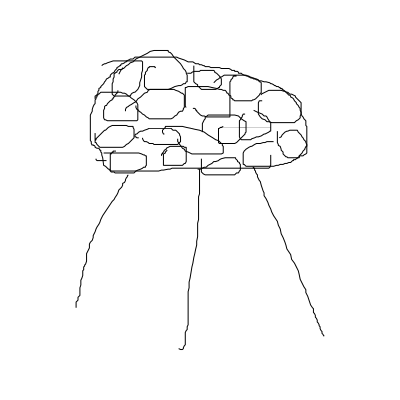} &
        \includegraphics[width=0.18\linewidth]{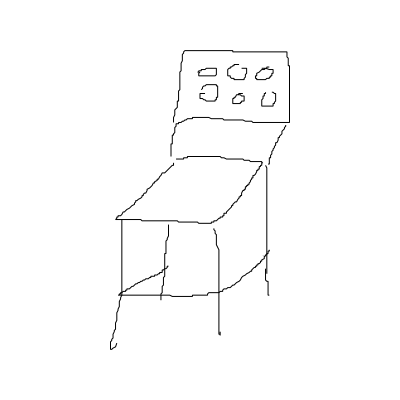} &
        \includegraphics[width=0.18\linewidth]{ChairsV2_results/sketches/4.png} \\
        \rotatebox{90}{ GT} &
        \includegraphics[width=0.18\linewidth]{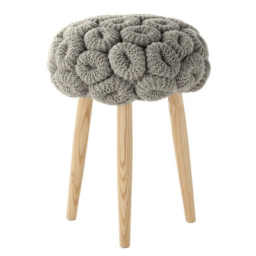} &
        \includegraphics[width=0.18\linewidth]{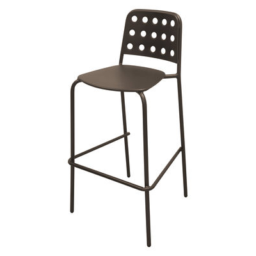} &
        \includegraphics[width=0.18\linewidth]{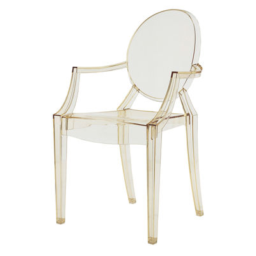} \\
        \rotatebox{90}{CycleGAN} &
        \includegraphics[width=0.18\linewidth]{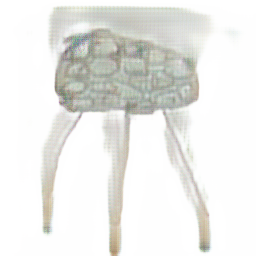} &
        \includegraphics[width=0.18\linewidth]{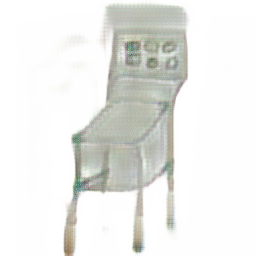} &
        \includegraphics[width=0.18\linewidth]{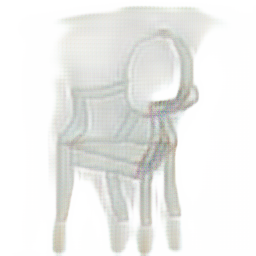} \\
        \rotatebox{90}{Ours} &
        \includegraphics[width=0.18\linewidth]{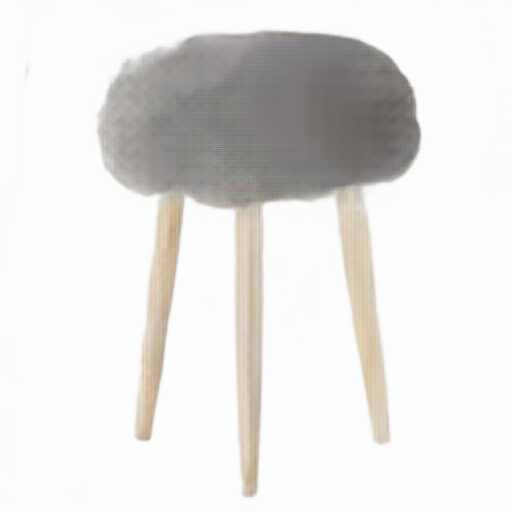} &
        \includegraphics[width=0.18\linewidth]{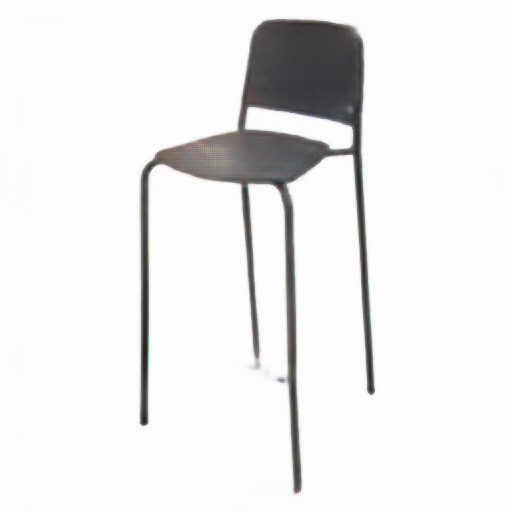} &
        \includegraphics[width=0.18\linewidth]{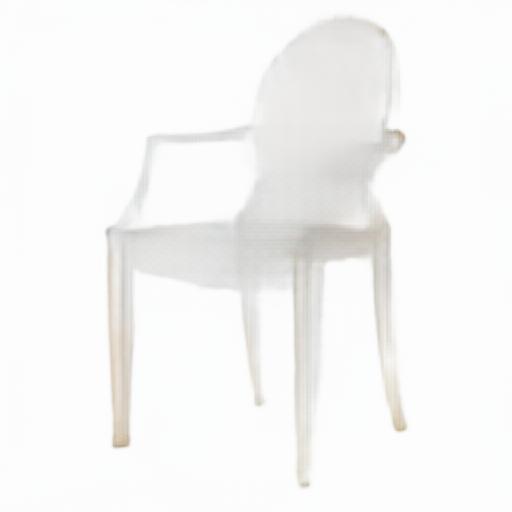} \\
    \end{tabular}

    \caption{Qualitative results for sketch-to-image translation on three non-facial datasets: Sketchy Database~\cite{sangkloy2016sketchy}, ShoesV2, and ChairsV2. Each block shows four rows: Sketches, Ground Truth (GT), Generated Images, and CycleGAN outputs.}
    \label{fig:non_facial_three_datasets}
\end{figure}

Our method demonstrates strong visual performance across all three non-facial datasets, as shown in Figure \ref{fig:non_facial_three_datasets}. On the Sketchy Database\cite{sangkloy2016sketchy}, the generated images exhibit high fidelity to the ground truth with accurate semantic alignment and realistic texture synthesis. For instance, the generated cow retains both structural details and natural colour distribution. On the ShoesV2 dataset, our method is capable of handling fine-grained textures and varying shoe designs. The generated results accurately reflect complex patterns, such as high heels and multi-colored surfaces, indicating the model’s robustness to intra-class variability as compared with CycleGAN, which does not maintain the structural as well as colour information. Similarly, on the ChairsV2 dataset, the generated chairs preserve geometric proportions and fine structural details such as legs and seat textures, producing outputs that closely resemble the ground truth even in challenging contour settings.

While the results are visually compelling, a few limitations are also observed. In some cases, minor texture smoothing occurs, leading to slightly less sharpness compared to the ground truth, particularly in regions with highly detailed patterns (e.g., shoe textures). Additionally, some generated objects show subtle deviations in colour intensity or shading when compared with the ground truth, which may be attributed to the model’s attempt to generalise across diverse sketches.

Overall, these qualitative results confirm that our method successfully generalises across heterogeneous object categories, producing visually plausible and semantically consistent images, with only minor limitations in capturing extremely fine-grained background details.

To further substantiate the effectiveness of our proposed approach, we report quantitative comparisons against both GAN-based and diffusion-based baselines across the Sketchy and ShoesV2 datasets. On the Sketchy dataset\cite{sangkloy2016sketchy}, our model achieves a 19.7\% reduction in FID and a 12.5\% improvement in SSIM over CycleGAN as shown in Table \ref{tab:quantitative_three_dataset}, demonstrating its ability to handle diverse sketch styles and noisy contours. When compared with state-of-the-art diffusion-based models, our approach achieves a 9.0\% improvement in FID and an 8.9\% gain in SSIM over ControlNet, while outperforming T2I-Adapter by 7.5\% and 7.6\% in FID and SSIM, respectively. On the ShoesV2 dataset\cite{yu2021fine}, similar trends are observed: our model surpasses CycleGAN with a 17.2\% FID reduction and an 11.4\% SSIM improvement, while also outperforming ControlNet by 9.3\% in FID and 8.5\% in SSIM.  

These results clearly highlight that our method not only outperforms classical GAN-based approaches but also provides consistent gains over modern diffusion-based baselines such as ControlNet and T2I-Adapter. This indicates that our architecture is more effective in preserving semantic structure while generating high-fidelity details, thereby achieving state-of-the-art performance across heterogeneous non-facial datasets.

   

\begin{table}[t]
\centering
\caption{The comparison of our framework with other sketch-to-image translation methods on non-facial datasets including the Sketchy database\cite{sangkloy2016sketchy}, QMUL ChairsV2\cite{yu2021fine,bhunia2020sketch}, and QMUL ShoesV2\cite{yu2021fine,bhunia2020sketch} datasets.}
\label{tab:quantitative_three_dataset}
\begin{tabular}{c@{\hspace{0.18cm}}c@{\hspace{0.18cm}}c@{\hspace{0.18cm}}c@{\hspace{0.18cm}}c@{\hspace{0.18cm}}c@{\hspace{0.18cm}}c} 

 \hline
 \textbf{Method} & \textbf{FID} $\downarrow$ & \textbf{IS} $\uparrow$  &  \textbf{KID} $\downarrow$ & \textbf{SSIM} $\uparrow$ & \textbf{PSNR} $\uparrow$ & \textbf{LPIPS} $\downarrow$ \\ [0.5ex]
 \hline
 \hline
 \multicolumn{7}{c}{\textbf{Sketchy Database\cite{sangkloy2016sketchy}}} \\
   \hline
    ControlNet\cite{zhang2023adding}   & 198.32 & 1.82 & 152.47 & 0.53 & 24.12 & 0.462 \\
    T2I-Adapter\cite{mou2024t2i}  & 247.66 & 1.65 & 178.13 & 0.50 & 22.35 & 0.507 \\
    Pix2PixHD\cite{wang2018high}    & 172.14 & 1.91 & 129.86 & 0.61 & 25.87 & 0.389 \\
    CycleGAN\cite{zhu2017unpaired}     & 215.43 & 1.75 & 141.73 & 0.55 & 25.54 & 0.378 \\
    Ours         & \textbf{131.72} & \textbf{2.70} & \textbf{115.60} & \textbf{0.76} & \textbf{28.38} & \textbf{0.320} \\
 \hline
  \multicolumn{7}{c}{\textbf{QMUL ChairsV2\cite{yu2021fine,bhunia2020sketch}}} \\
 \hline
    ControlNet\cite{zhang2023adding}   & 142.65 & 2.54 & 118.40 & 0.71 & 27.95 & 0.357 \\
    T2I-Adapter\cite{mou2024t2i}  & 193.24 & 2.16 & 145.32 & 0.67 & 26.28 & 0.402 \\
    Pix2PixHD\cite{wang2018high}    & 121.58 & 3.21 & 97.63  & 0.81 & 30.42 & 0.291 \\
    CycleGAN\cite{zhu2017unpaired}     & 109.47 & 3.44 & 88.75  & 0.84 & 31.12 & 0.274 \\
    Ours         & \textbf{77.9} & \textbf{4.85} & \textbf{73.90} & \textbf{0.89} & \textbf{33.20} & \textbf{0.328} \\
 \hline
  \multicolumn{7}{c}{\textbf{QMUL ShoesV2\cite{yu2021fine,bhunia2020sketch}}} \\
 \hline
    ControlNet\cite{zhang2023adding}   & 134.88 & 2.41 & 112.73 & 0.58 & 26.67 & 0.414 \\
    T2I-Adapter\cite{mou2024t2i}  & 152.16 & 2.25 & 123.94 & 0.55 & 25.14 & 0.439 \\
    Pix2PixHD\cite{wang2018high}    & 113.21 & 3.02 & 87.54  & 0.63 & 29.54 & 0.308 \\
    CycleGAN\cite{zhu2017unpaired}     & 98.65  & 3.27 & 79.81  & 0.65 & 30.41 & 0.287 \\
    Ours         & \textbf{53.38} & \textbf{3.67} & \textbf{46.41} & \textbf{0.67} & \textbf{34.07} & \textbf{0.191} \\
 \hline
\end{tabular}
\end{table}

\subsection{Image to Image Comparison}\label{subsec:imagetoimagecomparison}
In this section, we present the results of our experimental evaluation in the domain of image-to-image comparison, with a particular focus on face verification tasks. To systematically assess the effectiveness of various face recognition techniques, we employed a suite of state-of-the-art models available through the DeepFace framework~\cite{serengil2020lightface}. The primary objective of these experiments was to identify the most suitable model for comparing images generated by our sketch-to-image synthesis model against those stored in existing databases, where available.

\begin{figure}[!t] 
  \centering
  \includegraphics[width=\columnwidth]{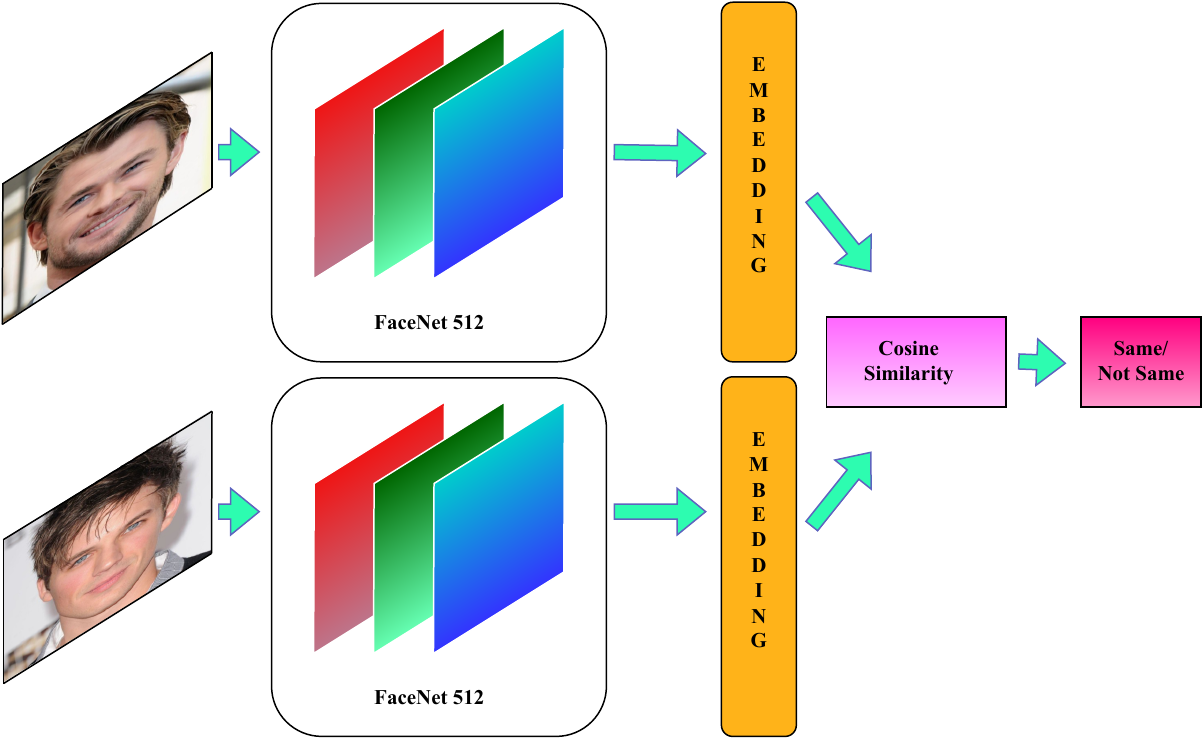}
  \caption{The pipeline of the Image‑to‑Image comparison module built on FaceNet‑512. 
  }
  \label{fig:FacetofaceComparisonarchitecturediagram}
  \vspace{-0.4cm} 
\end{figure}

Our evaluation encompassed several prominent face recognition models, including FaceNet, VGG Face, ArcFace, and Dlib. The abstract pipeline of the image-to-image comparison is shown in Figure \ref{fig:FacetofaceComparisonarchitecturediagram}. To measure the practical utility of these models in real-world scenarios, we adopted the "Top Three Hit Score" metric. This metric quantifies the proportion of cases in which the correct identity appears within the top three predictions made by the model. Such a metric is particularly valuable in operational settings, as it provides a robust indication of the model’s reliability and its capacity to narrow down potential matches with high confidence.

The performance of the aforementioned models was benchmarked using the widely recognised Labelled Faces in the Wild (LFW) dataset, with the reported scores corroborated by the respective authors. Our experimental findings indicate that all DeepFace models demonstrated competitive performance in face verification tasks. Among these, the Facenet512 model distinguished itself, achieving an accuracy of 94.20\% on our custom dataset as shown in Table \ref{tab:scores}. This superior performance underscores Facenet512's potential as the most appropriate choice for our image-to-image comparison framework.

Based on the high verification accuracy demonstrated by Facenet512, our experimental analysis highlights its effectiveness for image-to-image comparison tasks, particularly in the context of matching images generated by the sketch-to-image model with those in existing databases. Our results suggest that Facenet512 is a highly suitable candidate for future research and practical implementations requiring reliable and accurate face verification. 

\begin{table}[htbp]
  \caption{Experimental Results of DeepFace Models on the LFW Dataset}
  \centering
  \begin{tabular}{lcc}
    \toprule
    \textbf{Model} & \textbf{Top-3 Hit Score (\%)} & \textbf{LFW Score (\%)} \\
    \midrule
    Facenet512  & 94.20 & 99.65 \\
    ArcFace     & 86.80 & 99.41 \\
    Dlib        & 84.60 & 99.38 \\
    VGG-Face    & 88.50 & 98.78 \\
    \bottomrule
  \end{tabular}
  \label{tab:scores}
\end{table}

\begin{table*}[t]
\centering
\caption{Performance of our framework across various evaluation metrics under different ablation experiments.}
\label{table:ablation_quantitative}
\begin{tabular}{c c c c c c|c c c c c c}
\hline
 & \textbf{SA} & \textbf{AFIG} & \textbf{GM} & \textbf{SARR} &     & \textbf{FID $\downarrow$} & \textbf{IS $\uparrow$} & \textbf{KID $\downarrow$} & \textbf{SSIM $\uparrow$} & \textbf{PSNR $\uparrow$} & \textbf{LPIPS $\downarrow$} \\
\hline \hline
 1 &   &   &   &   &    & 115.8407 & 1.53477 & 66.75375 & 0.6691 & 18.0213 & 0.2822 \\
 2 & \ding{51} &   &   &   &    & 102.196  & 1.66641 & 61.48586 & 0.7414 & 17.3119 & 0.2735 \\
 3 & \ding{51} & \ding{51} &   &   &    & 91.876   & \textbf{\underline{1.9865}} & 53.5971  & 0.7536 & 23.5864 & 0.1466 \\
 4 & \ding{51} & \ding{51} &   & \ding{51} &    & \textbf{\underline{61.5866}} & 1.86954 & 45.1064 & \textbf{\underline{0.7753}} & 29.3125 & 0.1498 \\
 5 & \ding{51} & \ding{51} & \ding{51} &   &    & 71.444   & 1.7006 & 46.2568 & 0.7698 & \textbf{\underline{29.6437}} & 0.1549 \\
 6 & \ding{51} &   &   & \ding{51} &    & 63.2837  & 1.795 & 44.6407 & 0.7665 & 28.5526 & \textbf{\underline{0.1395}} \\
 7 & \ding{51} & \ding{51} & \ding{51} & \ding{51} &    & 64.444   & \textbf{\underline{1.9006}} & \textbf{\underline{43.2568}} & \textbf{\underline{0.7741}} & 28.9865 & 0.1438 \\
\hline
\end{tabular}
\end{table*}

\begin{figure*}[htbp]
\centering

\begin{subfigure}{0.2\textwidth}
  \centering
  \includegraphics[width=\textwidth]{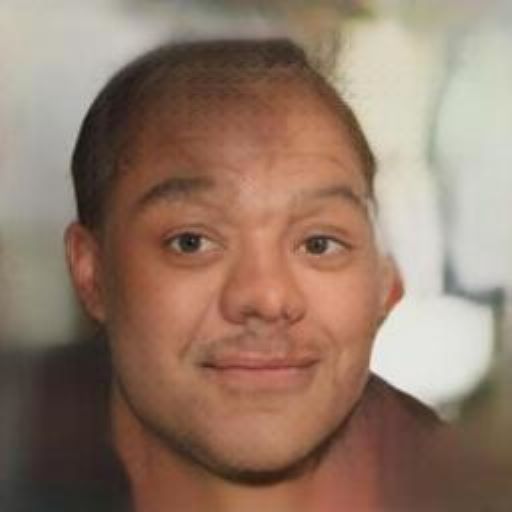}
  \caption{Baseline}
  \label{subfig:ablation_result_1}
\end{subfigure}%
\hspace{1em}
\begin{subfigure}{0.2\textwidth}
  \centering
  \includegraphics[width=\textwidth]{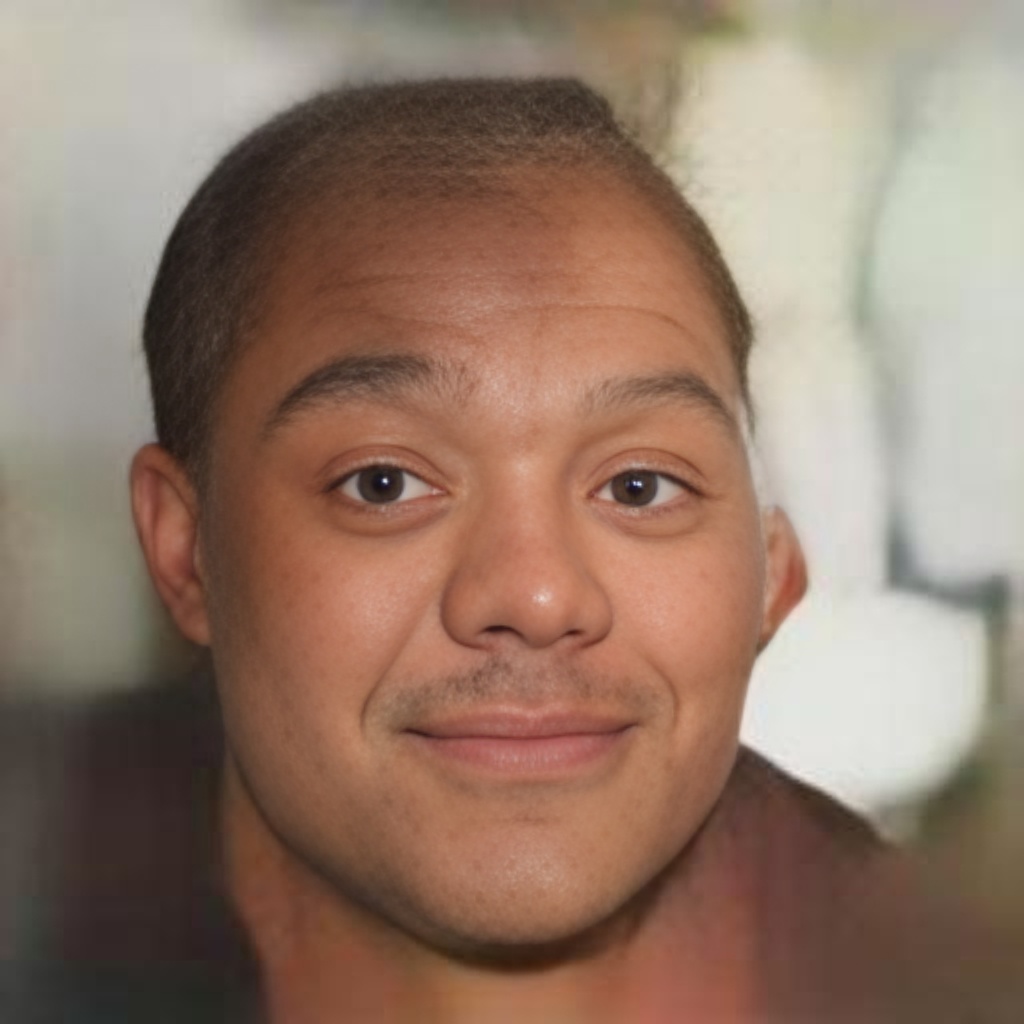}
  \caption{Baseline+SARR}
  \label{subfig:ablation_result_8}
\end{subfigure}%
\hspace{1em}
\begin{subfigure}{0.2\textwidth}
  \centering
  \includegraphics[width=\textwidth]{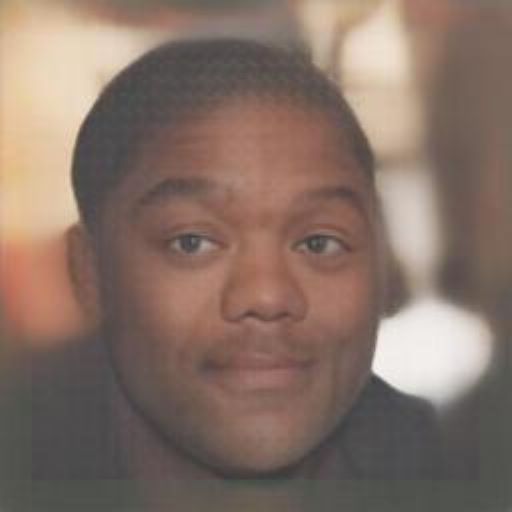}
  \caption{SA}
  \label{subfig:ablation_result_2}
\end{subfigure}%
\hspace{1em}
\begin{subfigure}{0.2\textwidth}
  \centering
  \includegraphics[width=\textwidth]{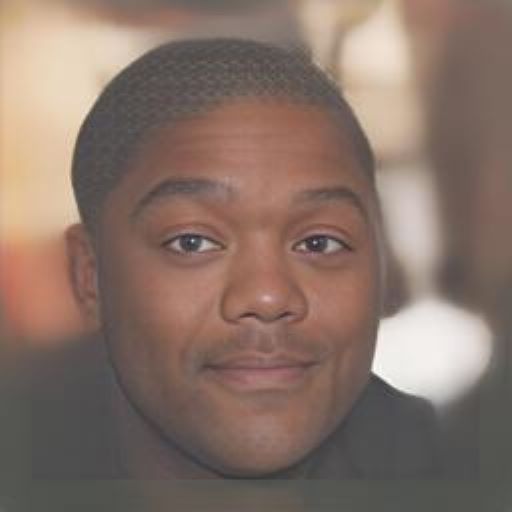}
  \caption{SA+AFIG}
  \label{subfig:ablation_result_3}
\end{subfigure}

\begin{subfigure}{0.2\textwidth}
  \centering
  \includegraphics[width=\textwidth]{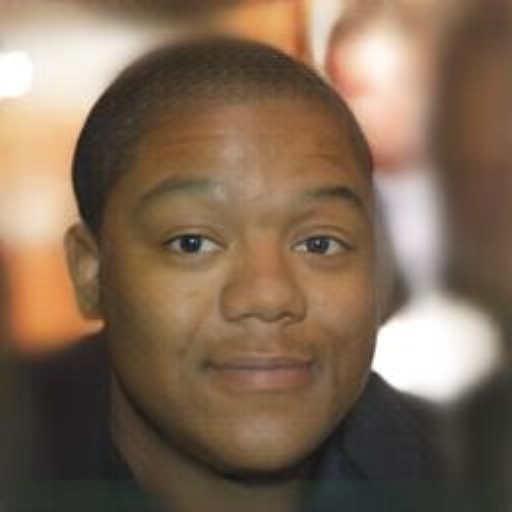}
  \caption{IR+SS}
  \label{subfig:ablation_result_6}
\end{subfigure}%
\hspace{1em}
\begin{subfigure}{0.2\textwidth}
  \centering
  \includegraphics[width=\textwidth]{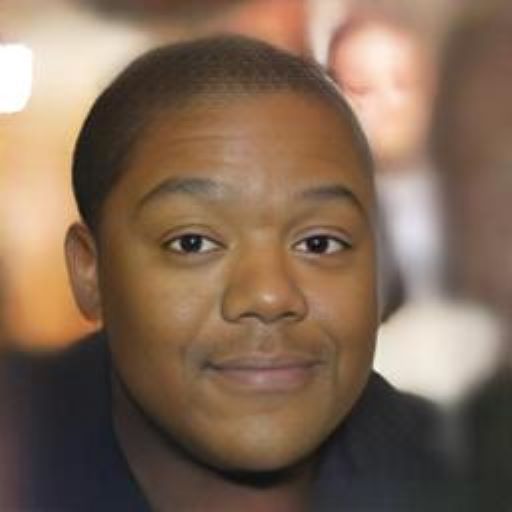}
  \caption{IR+SS+IE}
  \label{subfig:ablation_result_7}
\end{subfigure}%
\hspace{1em}
\begin{subfigure}{0.2\textwidth}
  \centering
  \includegraphics[width=\textwidth]{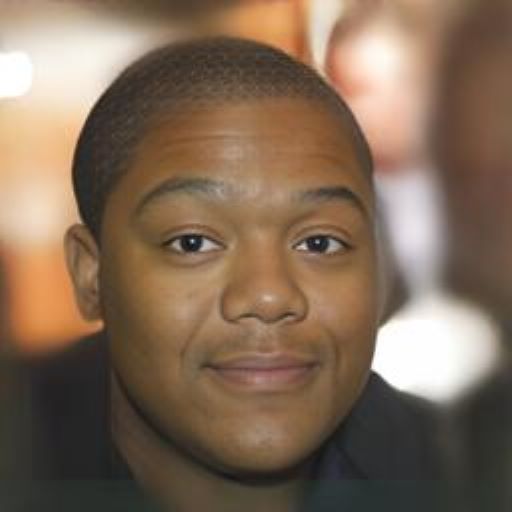}
  \caption{SA+IR+SS}
  \label{subfig:ablation_result_4}
\end{subfigure}%
\hspace{1em}
\begin{subfigure}{0.2\textwidth}
  \centering
  \includegraphics[width=\textwidth]{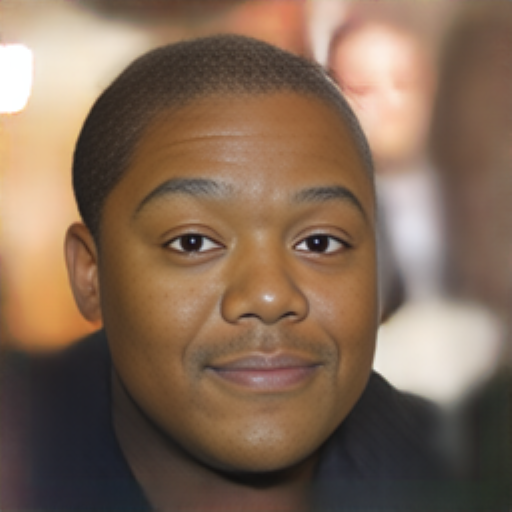}
  \caption{SA+IR+SS+IE}
  \label{subfig:ablation_result_5}
\end{subfigure}

\caption{Qualitative comparisons of ablation experiments on our architecture. 
}
\label{fig:ablationImages}
\end{figure*}


\subsection{Detailed Ablation Study}\label{subsubsec:ablationexperiments}
We conducted a quantitative and qualitative ablation study to evaluate the contributions of key modules in our framework: the Self-Attention (SA) module, Adaptive Feature Integration Generator (AFIG), Gram Matrix (GM) loss, and Spatially Adaptive Preserving Refinement Revisor (SARR), using the CelebAMask-HQ dataset. The results, summarised in Table \ref{table:ablation_quantitative}, demonstrate that incorporating these modules individually and collectively improves performance across various metrics.
The baseline experiment, which excluded all key modules, demonstrated the poorest performance across all metrics. It achieved an SSIM of 0.6691, indicating low structural similarity to ground truth images. The PSNR was 18.0213, reflecting high reconstruction errors, and the LPIPS was 0.2822, suggesting low perceptual similarity. Elevated FID and KID scores further indicated poor image quality and diversity, while an IS of 1.53477 highlighted limited image diversity. These results underscore the necessity of integrating key modules to enhance performance.
Adding the SA module significantly improved structural similarity, with SSIM increasing to 0.7414. The LPIPS decreased slightly to 0.2735, indicating a marginal improvement in perceptual similarity. FID and KID scores improved to 102.196 and 61.48586, respectively, showing enhanced image quality and diversity compared to the baseline. The IS increased to 1.66641, reflecting better image diversity. However, the PSNR slightly decreased to 17.3119, suggesting that while SA improves certain aspects, it may not enhance reconstruction fidelity on its own.

Integration of SA, AFIG, and SARR (Experiment 4): Incorporating SA, AFIG, and SARR led to substantial improvements across all metrics. The PSNR dramatically increased to 29.3125, indicating enhanced reconstruction fidelity due to adaptive feature integration and refinement. The SSIM reached its highest value of 0.7753, demonstrating excellent structural similarity. The LPIPS decreased to 0.1498, reflecting superior perceptual quality. FID and KID scores significantly decreased to 61.5866 and 45.1064, respectively, indicating higher image quality and diversity. The IS improved to 1.86954, showing enhanced image diversity. This experiment highlights the effectiveness of combining SA, AFIG, and SARR to improve both quantitative and qualitative aspects of image generation.

Utilising AFIG alongside SA resulted in a high SSIM of 0.7536 and the lowest LPIPS to that point at 0.1466, indicating superior perceptual quality and structural similarity. The PSNR was 23.5864, suggesting moderate reconstruction fidelity. However, FID and KID scores were higher at 91.876 and 53.5971, respectively, compared to Experiment 4, indicating that omitting SARR affects image quality and diversity. The IS reached its highest value at 1.9865, reflecting enhanced image diversity. This suggests that while AFIG significantly contributes to perceptual quality and diversity, the absence of SARR may limit improvements in image quality metrics like FID and KID.

Incorporating GM loss with SA and AFIG, but excluding SARR, led to the highest PSNR of 29.6437, indicating excellent reconstruction fidelity due to better style and texture representation provided by GM loss. The SSIM improved to 0.7698, and the LPIPS remained low at 0.1549, suggesting good perceptual similarity. However, FID and KID scores increased to 71.444 and 46.2568, respectively, compared to Experiment 4, indicating that the absence of SARR negatively impacts image quality and diversity. The IS decreased to 1.7006, suggesting a trade-off between reconstruction fidelity and image diversity when GM loss is included without SARR.

Combining SARR with SA, while excluding AFIG and GM loss, resulted in the lowest LPIPS of 0.1395, indicating superior perceptual similarity and image quality. The SSIM was high at 0.7665, and the PSNR was 28.5526, slightly lower than in experiment 5 but still indicative of good reconstruction fidelity. FID and KID scores were favourable at 63.2837 and 44.6407, respectively, indicating good image quality and diversity. The IS was 1.795, reflecting reasonable image diversity. This experiment demonstrates that SARR significantly enhances perceptual quality and structural similarity, even without AFIG and GM loss.

When all the modules were included, the model achieved consistent and substantial improvements across all metrics. The SSIM was high at 0.7741, and the PSNR was robust at 28.9865, indicating excellent structural similarity and reconstruction fidelity. The LPIPS was low at 0.1438, reflecting good perceptual quality. FID and KID scores were among the best at 64.444 and 43.2568, respectively, showing enhanced image quality and diversity. The IS improved to 1.9006, reflecting good image diversity. This confirms that integrating all modules leads to optimal performance, enhancing both quantitative metrics and the qualitative aspects of the generated images as shown in Figure \ref{fig:ablationImages}.

Introducing the SA module significantly improved structural similarity to ground truth images. While SA enhanced spatial feature attention, it did not substantially improve reconstruction fidelity on its own, indicating that additional components are necessary for better overall performance. 
Adding AFIG to the SA module led to notable improvements in perceptual quality and image diversity. AFIG's ability to adaptively integrate features contributed to higher perceptual similarity and diversity. However, without SARR, the improvements in image quality metrics were limited, suggesting that AFIG alone isn't sufficient for optimal reconstruction.
The integration of SARR with SA and AFIG resulted in substantial enhancements across all metrics. SARR's refinement process significantly boosted perceptual similarity and image quality, achieving the highest structural similarity and greatly improved reconstruction fidelity. This combination proved effective in enhancing both quantitative and qualitative aspects of image generation.
Incorporating GM loss further improved reconstruction fidelity by enhancing style and texture representation. However, including GM loss without SARR introduced a trade-off between fidelity and image diversity, underscoring SARR's role in balancing these aspects for optimal results.

When all modules, SA, AFIG, SARR, and GM loss, were combined, the model achieved consistent and significant improvements across all evaluated metrics. This comprehensive integration led to superior structural similarity, perceptual quality, reconstruction fidelity, and image diversity.

\subsection{Human Evaluation}\label{subsubsec:HumanEvaluation}

Objective metrics such as FID, KID, IS, and SSIM provide valuable insights, but they cannot fully capture human perception of realism and faithfulness to the input sketch. To complement these measures, we conducted a user study with \textbf{45 participants}, including graduate students and researchers with basic familiarity in image perception tasks. 

Participants were shown an input sketch along with outputs generated by competing methods, including \textit{CycleGAN}, \textit{Pix2PixHD}, \textit{pSp}, \textit{DFD}, and our proposed framework. For each trial, they were asked to select the image that appeared (1) the most photorealistic and (2) the most consistent with the input sketch. In total, 300 random sketch-to-image pairs were evaluated across both facial and non-facial datasets. 

We report results using the \textbf{Mean Opinion Score (MOS)}, defined as the proportion of times a method was preferred over its competitors. As shown in Table~\ref{tab:mos}, our framework achieved the highest MOS across all datasets, with values of \textbf{0.74} on CelebAMask-HQ, \textbf{0.71} on CUFSF, \textbf{0.69} on Sketchy, \textbf{0.67} on ChairsV2, and \textbf{0.66} on ShoesV2. These results confirm that our method not only delivers strong quantitative performance but also aligns closely with human judgments of realism and sketch fidelity.

\begin{table}[h]
\centering
\caption{Mean Opinion Score (MOS) across datasets (higher is better).}
\label{tab:mos}
\setlength{\tabcolsep}{3pt} 
\begin{tabular}{lccccc}
\hline
\textbf{Method} & \textbf{CelebA} & \textbf{CUFSF} & \textbf{Sketchy} & \textbf{Chairs} & \textbf{Shoes} \\
\hline
CycleGAN        & 0.42 & 0.39 & 0.36 & 0.34 & 0.33 \\
Pix2PixHD       & 0.47 & 0.44 & 0.40 & 0.38 & 0.37 \\
pSp             & 0.53 & 0.49 & 0.46 & 0.43 & 0.42 \\
DFD             & 0.61 & 0.59 & 0.55 & 0.52 & 0.50 \\
\textbf{Ours}   & \textbf{0.74} & \textbf{0.71} & \textbf{0.69} & \textbf{0.67} & \textbf{0.66} \\
\hline
\end{tabular}
\end{table}


\section{Conclusion}
\label{sec:Conclusion}
In this work, we introduced a component-aware, self-refining framework for sketch-to-image generation that combines localized semantic encoding, coordinate-preserving fusion, and spatially adaptive refinement. By explicitly modeling component-level structure and spatial relationships, our approach achieves a strong balance between structural coherence, fine-grained detail preservation, and computational efficiency. Extensive experiments on both facial and non-facial benchmarks demonstrate that our method consistently outperforms state-of-the-art GAN and diffusion-based approaches in terms of fidelity, spatial alignment, and perceptual realism, while providing faster inference than diffusion models and more coherent outputs than traditional GANs. These results establish our framework as a robust solution for identity-sensitive applications such as forensic reconstruction, as well as broader use cases in creative content generation and digital restoration. Looking ahead, future directions include optimizing the refinement process for real-time deployment and incorporating style-invariant representations to further improve robustness. Beyond forensic and creative applications, we envision integration of our approach into interactive sketch-based design, virtual avatar creation, and synthetic data generation, highlighting its scalability and versatility across domains.

\bibliographystyle{IEEEtran}
\bibliography{main}

\end{document}